\documentclass[sigconf,nonacm]{acmart}

\usepackage{enumitem}

\usepackage{booktabs}
\usepackage{xcolor}
\usepackage{colortbl}
\usepackage{etoolbox}
\usepackage{tcolorbox}\tcbuselibrary{skins,breakable}
\usepackage{pifont}
\usepackage{soul}
\usepackage{placeins}
\usepackage{etoc}
\usepackage{flushend}
\definecolor{tabhead}{HTML}{DFE9F6}
\definecolor{tabours}{HTML}{EEF5F1}
\definecolor{tabzebra}{HTML}{F5F7FA}
\newcommand{\headrow}{\rowcolor{tabhead}}
\newcommand{\zebrarow}{\rowcolor{tabzebra}}
\newcommand{\oursrow}{\rowcolor{tabours}}
\definecolor{linkblue}{HTML}{184694}
\AtEndPreamble{\hypersetup{colorlinks=true,linkcolor=linkblue,citecolor=linkblue,urlcolor=linkblue,filecolor=linkblue}}
\newcommand{\headrulebg}{\noalign{{\color{tabhead}\hrule height\cmidrulewidth}\vskip-\cmidrulewidth}}
\AtBeginEnvironment{table}{}
\AtBeginEnvironment{table*}{}

\setcopyright{none}
\renewcommand\footnotetextcopyrightpermission[1]{}

\AtBeginDocument{%
  }

\setcopyright{acmlicensed}
\copyrightyear{2027}
\acmYear{2027}
\acmDOI{}
\acmConference[]{}
\acmISBN{}

\begin{document}

\title[AskGLP1]{AskGLP1: A Knowledge Graph LLM Agent for Exploring GLP-1 Experience from Online Patient Narratives}

\author{
  Zhaoqian Xue\textsuperscript{1,\textdagger},
  Anay Somani\textsuperscript{1,*},
  Siddhartha Singh\textsuperscript{1,*},
  Junting Mu\textsuperscript{2},
  Zhongyin Wang\textsuperscript{3},\\
  Haozheng~Liu\textsuperscript{4},
  Ayesha Khatun\textsuperscript{1},
  Yeaeun Kwon\textsuperscript{1},
  Renkai Ma\textsuperscript{5},\\
  Siyuan Ma\textsuperscript{6},
  Zhe He\textsuperscript{7},
  Danxia Yu\textsuperscript{6},
  Lingyao Li\textsuperscript{1,\textdagger}
  \\
  \mbox{\textsuperscript{1}University of Arizona} \quad
  \mbox{\textsuperscript{2}University of California, Irvine} \quad
  \mbox{\textsuperscript{3}University of Illinois Urbana-Champaign} \quad
  \mbox{\textsuperscript{4}New York University} \quad
  \mbox{\textsuperscript{5}University of Cincinnati} \quad
  \mbox{\textsuperscript{6}Vanderbilt University Medical Center} \quad
  \mbox{\textsuperscript{7}Florida State University}
}

\begin{abstract}
  GLP-1 medications are increasingly used for weight management, and many people share their experiences online. These narratives show why people take the drugs and what they experience, but they remain unstructured and disconnected from formal pharmacovigilance. We therefore present AskGLP1, a system that transforms online patient narratives into a knowledge graph (KG) and answers questions about medication experiences. Through an LLM-assisted pipeline grounded in the MedDRA terminology, AskGLP1 screens nearly 8.5 million Reddit posts for medication experiences and distills 163,135 of them, with 2,506 WebMD reviews, into a KG of 385,790 drug--side-effect and drug--purpose relations. We find that GLP-1 experiences are source- and drug-specific. Online narratives share the most common gastrointestinal symptoms with FAERS, but they foreground everyday burdens such as fatigue and food aversion, whereas FAERS reports more often include injection-site reactions and selected serious diagnoses such as pancreatitis. We further find that posts report less nausea and vomiting for tirzepatide than for semaglutide, more fatigue for Wegovy than Ozempic and for Zepbound than Mounjaro. Evaluations show that the agent answers with verifiable evidence and declines requests for medical advice. Our findings reveal gaps between what FAERS captures and what people report online, and they call for patient-centered approaches to pharmacovigilance. Code, prompts, and evaluation questions are available at \url{https://github.com/ZhaoqianXue/AskGLP1}.
\end{abstract}




\renewcommand{\shortauthors}{Xue et al.}

\maketitle
{\renewcommand{\thefootnote}{}%
\footnotetext{\textsuperscript{*}Equal contribution as co-second authors.\newline\textsuperscript{\textdagger}Corresponding authors: zhaoqianxue@arizona.edu, lingyaoli@arizona.edu}}

\vspace{-2mm}
\section{Introduction}
\label{sec:intro}

GLP-1 medications, a term we use for glucagon-like peptide-1 (GLP-1) receptor agonists such as semaglutide and for the dual GIP/GLP-1 receptor agonist tirzepatide, were originally developed for type 2 diabetes and are now widely prescribed for weight management~\cite{wilding2021once,jastreboff2022tirzepatide}. By late 2025, 12\% of U.S. adults reported taking one~\cite{kff2025glp1}, but side effects are common. Nausea and diarrhea are the most frequent in trials~\cite{wilding2021once, ismaiel2025gastrointestinal}, and observational studies link weight-loss use to rarer serious outcomes such as pancreatitis, bowel obstruction, and gastroparesis~\cite{sodhi2023risk}. Understanding these patterns requires knowing why people take these drugs and what they experience.

Conventional data sources capture these experiences only partially~\cite{hazell2006under}. The FDA Adverse Event Reporting System (FAERS) collects coded adverse event and medication error reports, but most of its reports are submitted by manufacturers and tend to emphasize serious outcomes, while spontaneous reporting misses many adverse events~\cite{moore2016completeness}. By contrast, many people describe them in detail on Reddit and WebMD. Such posts can extend FAERS signals~\cite{park2022comparison}, and recent work examines GLP-1 discussions on Facebook and Reddit~\cite{alibilli2025harnessing,sehgal2026self, duan2026crowdsourcing}. These communities thus offer a large, informal evidence source that complements formal pharmacovigilance.

Turning these posts into reliable evidence is challenging. First, many symptom mentions do not report a medication experience, because people also ask questions, give advice, or share news~\cite{klein2017detecting,lamb-etal-2013-separating}. Second, online language is informal and unstructured, which needs to be mapped to standard terms such as MedDRA for cross-source comparison~\cite{fan-etal-2025-llm,park2022comparison}. Third, the evidence should be accessible and verifiable. General-purpose large language model (LLM) chatbots can give unsupported or hallucinated health answers~\cite{zhu2025can, pandit2025medhallu}. Answers should therefore cite their sources and flag missing evidence.

Recent studies show promise. One study uses LLMs to build a knowledge graph (KG) of semaglutide side effects from Reddit~\cite{duan2026crowdsourcing}, though it does not support natural language queries. LLM agents for medical question answering integrate KGs or evidence retrieval~\cite{rezaei-etal-2025-agentic,jia2026agentic}, drawing on curated clinical knowledge rather than patient narratives. Other agents improve traceability by grounding answers in source-linked evidence~\cite{wang2025lins,yu2026knowledge}. Building on these approaches, we present \textbf{AskGLP1}, a system that organizes evidence from Reddit, WebMD, and FAERS into a KG and answers questions about the reasons for taking GLP-1 medications and reported side effects through a KG-grounded agent. We study three research questions:

\begin{itemize}[leftmargin=*, nosep]
  \item \textbf{RQ1:} What side effects do people report online, and how do reporting patterns vary across drugs and change over time?
  \item \textbf{RQ2:} How do online patient narratives complement FAERS?
  \item \textbf{RQ3:} How does the KG-grounded agent provide verifiable answers to AskGLP1 users' questions?
\end{itemize}

To answer these questions, AskGLP1 combines an LLM pipeline that converts posts into a quote-grounded KG with a KG-grounded agent that answers questions over it. The pipeline detects medication experiences, extracts drug--purpose and drug--side-effect relations, links each to its sources, and normalizes side effects to MedDRA while keeping users' wording. The agent queries the KG, returns supporting quotes, and abstains when evidence is missing. We make four contributions. First, we build a quote-grounded KG covering multiple GLP-1 medications and purposes. Second, we analyze side effects reported online to show how reporting patterns vary across drugs and change over time. Third, we compare Reddit, WebMD, and FAERS to show where they agree and where patient narratives add to FAERS. Fourth, we evaluate AskGLP1 against baselines to show that its verifiable answers reach the correct conclusion far more often. Overall, we turn patient narratives on social media and the Web into verifiable evidence on real-world medication experiences for researchers, clinicians, and patients.
\vspace{-3mm}
\section{Related Work}
\label{sec:related}

\noindent\textbf{Patient-Reported Medication Experiences and FAERS.}
People often share their experiences with drugs online. These posts help researchers study why people take drugs, whether they find them helpful, and what side effects they report \cite{leaman-etal-2010-towards,frost2011patient}. These online experiences offer a patient perspective alongside adverse event reports in FAERS. Comparisons with FAERS reports find similar broad reporting patterns on Twitter and show that WebMD reviews can add personal details on side effects \cite{freifeld2014digital,park2022comparison}. How GLP-1 experiences on Reddit and WebMD differ from FAERS reports remains an open research question. Informal language and differences across websites and evaluation methods also make online posts hard to analyze \cite{sarker2015utilizing,grasser2018aspect}. More recently, a Facebook study examines GLP-1 symptom mentions, including personal experiences and general concerns~\cite{alibilli2025harnessing}. Sehgal et al. screen for personal semaglutide and tirzepatide use on Reddit and analyze reported side effects with manually evaluated classifiers~\cite{sehgal2026self}. They also report descriptive drug comparisons and monthly symptom counts.

\par\vspace{1mm}\noindent\textbf{LLM Extraction and Knowledge Graphs from User-Generated Content.}
Building a KG from patient narratives requires finding health information, understanding its meaning, and matching different entities for the same concept. Early studies extract health concepts and their relationships from Reddit. HEAL also organizes personal experiences into a KG \cite{foufi2019mining, welivita2022heal}. However, a symptom mention may describe something a person denies or worries about, rather than something they experienced \cite{scaboro2022increasing}. Later studies use LLMs to extract side effects and match everyday descriptions to standard medical terms \cite{berkowitz-etal-2024-tlab, fan-etal-2025-llm, naskar-etal-2025-hybrid}. 
More closely related to our work, a semaglutide study uses an LLM to build a KG of side effects from Reddit posts \cite{duan2026crowdsourcing}. Yu et al. combine Reddit, WebMD, and FAERS in a traceable KG with MedDRA mappings and an agent that queries it for antidepressant information~\cite{yu2026knowledge}.

\par\vspace{1mm} \noindent\textbf{LLM Agents for Evidence-Grounded Health Question Answering.}
Early research shows that LLMs hold useful medical knowledge, but their answers can still be wrong \cite{singhal2023large}. Retrieval-augmented generation (RAG) helps address this problem by adding retrieved information, although finding the right evidence remains important \cite{xiong-etal-2024-benchmarking, liu2025improving}. Agent systems extend this approach by managing searches and choosing evidence. For example, LINS provides source-linked medical answers, while an agent framework answers patient questions using selected sentences from health records \cite{wang2025lins, hwang-etal-2025-dmis}. Other work adds memory or medical KGs to support answers across several steps \cite{jia2026agentic, rezaei-etal-2025-agentic}. 
These findings guide AskGLP1’s KG-grounded agent, which can help AskGLP1 users explore people’s reported experiences with drugs across Reddit and WebMD and check the original text behind each answer.


\par\vspace{1mm}\noindent\textbf{Research Gaps.}
We extend prior GLP-1 studies by comparing ingredients, brands, and years while checking sensitivity to the basis of drug identification (RQ1). AskGLP1 connects relation-level evidence and MedDRA mappings to cross-source comparisons under explicit counting rules with a separate denominator for each source (RQ2). We evaluate the agent against two baselines through human ratings on 120 questions, assessing evidence support separately from major errors, including those involving counting scope (RQ3).

\vspace{-2mm}
\section{AskGLP1 System Design}
\label{sec:system}

\begin{figure*}[t]
  \centering
  \captionsetup{skip=5pt}
  \includegraphics[width=\textwidth]{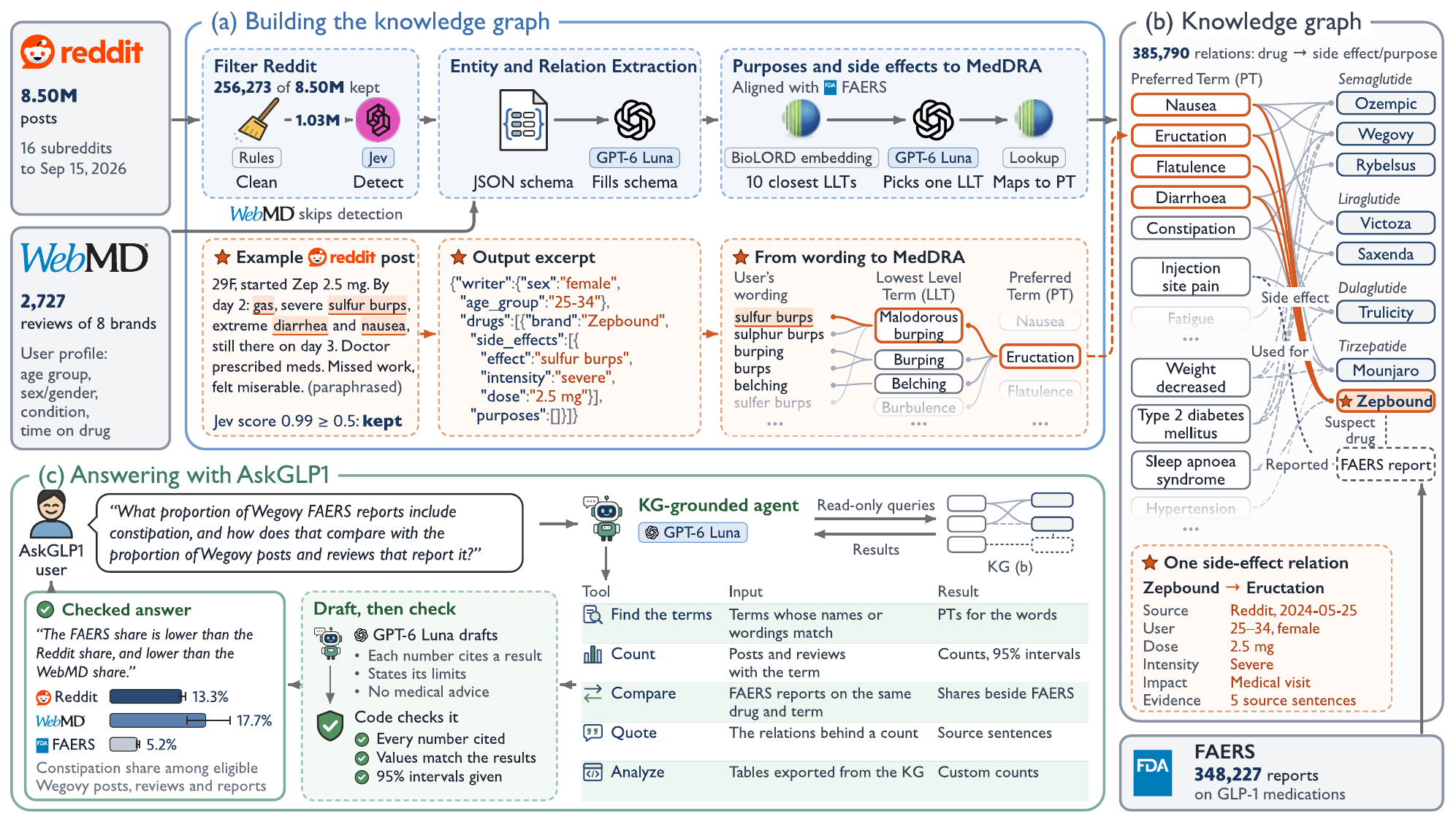}
  \caption{Overview of AskGLP1. (a) Reddit posts and WebMD reviews yield drug--side-effect and drug--purpose relations, with purposes and side effects coded to MedDRA. (b) The KG links drugs to purposes and side effects and connects FAERS reports to shared drug and MedDRA PT nodes. (c) The KG-grounded agent answers a question about constipation shares across sources for Wegovy. Denominators require a retained side effect or event. Numbers and quotes are checked against tool outputs.}
  \Description{Three panels show narrative extraction and MedDRA coding, a KG linked to FAERS reports, and an agent answer with checked numbers.}
  \label{fig:overview}
  \vspace{-4mm}
\end{figure*}

Figure~\ref{fig:overview} provides an overview of AskGLP1 system design. AskGLP1 builds a KG from Reddit and WebMD narratives about GLP-1 medications, then compares the patterns with FAERS, and allows users to query it in natural language (Table~\ref{tab:data}). After data collection (Section~\ref{sec:data}), AskGLP1 contains the following components.

\begin{itemize}[leftmargin=*, nosep]
\item \textbf{Medication experience detection} (Section~\ref{sec:detection}) identifies posts that report personal experience with details.
\item \textbf{Entity and relation extraction} (Section~\ref{sec:extraction}) turns each post into drug--purpose and drug--side-effect relations with details and demographics, each citing its sources.
\item \textbf{Entity normalization} (Section~\ref{sec:normalization}) maps drug names to a shared catalog and purposes and side effects to MedDRA terms, so that variants count as one concept and match FAERS.
\item \textbf{KG construction} (Section~\ref{sec:kg}) links the relations through extracted entities, so that the KG can be filtered by drug, source, date, and demographics and traced to its text.
\item \textbf{KG-grounded agent} (Section~\ref{sec:kgqa}) answers questions only from graph tools and checks each number and quote, so that AskGLP1 users can verify every claim.
\end{itemize}

\begin{table}[t]
\centering
\caption{AskGLP1 data preparation by step.}
\vspace{-2mm}
\label{tab:data}
\small
\begin{tabular}{lrrr}
\toprule
\headrow  & Reddit & WebMD & FAERS \\
\midrule
\zebrarow Collected & 8,496,004 & 2,727 & 397,362 \\
Cleaned & 1,030,309 & 2,681 & -- \\
\zebrarow Kept by detection & 256,273 & -- & -- \\
Eligible reports & -- & -- & 348,227 \\
\zebrarow Posts in the KG & 163,135 & 2,506 & -- \\
\quad Drug--Side-effect relations & 334,201 & 5,166 & -- \\
\zebrarow \quad Drug--Purpose relations & 43,875 & 2,548 & -- \\
\quad Age group stated & 15,129 & 2,413 & -- \\
\zebrarow \quad Sex/gender stated & 11,993 & 2,321 & -- \\
\bottomrule
\end{tabular}
\vspace{-5mm}
\end{table}

\vspace{-3mm}
\subsection{Data Collection and Preparation}
\label{sec:data}

We study four ingredients of GLP-1 medications and their eight brands, namely semaglutide (Ozempic, Wegovy, Rybelsus), tir\-ze\-pa\-tide (Mounjaro, Zepbound), liraglutide (Victoza, Saxenda), and dulaglutide (Trulicity). We select this set based on the availability of WebMD brand reviews and corresponding Reddit coverage to support comparisons across ingredients, brands, and sources. Together, these brands span approved indications for type 2 diabetes and weight management~\cite{fda2023glp1approvals,fda2023zepbound}.

\textbf{Reddit.} Reddit is a discussion platform organized into topic-specific forums (subreddits), where people publish submissions and reply in threaded comments~\cite{foufi2019mining}. Through the Arctic Shift archive~\cite{heitmann2026arcticshift}, we collect submissions and comments, together called \emph{posts} (as are WebMD reviews), from 16 GLP-1 and type 2 diabetes subreddits, each from the earliest FDA approval among the brands assigned to it in our collection configuration~\cite{fda2023glp1approvals,fda2023zepbound} to September 15, 2026 (Appendix~\ref{app:reproducibility}). Because some of these subreddits center on one brand or ingredient, which ties a post to an ingredient even when the user only says ``my shot,'' we apply no drug-name filter. This yields 8,496,004 posts (551,694 submissions and 7,944,310 comments), counted per post because we collect no account identifiers. Next, we apply rule-based cleaning to remove short, duplicate, templated, and non-English posts, resulting in a dataset of 4,660,528 posts. To give extraction more context, we retain posts with more than 40 distinct content words outside sentences repeated across posts. We fix this threshold before any LLM step because it yields the corpus size closest to the budgeted target of one million posts. This leaves 1,030,309 posts (Appendix~\ref{app:reproducibility}).

\textbf{WebMD.} WebMD is a consumer health website whose drug pages let people rate a drug and describe their experience with it~\cite{park2022comparison}. Each review concerns a named drug and carries profile fields entered by the user, such as age group, sex/gender, and the condition treated. We collect all 2,727 reviews of the eight brands posted up to September 2026, of which 2,681 contain text.

\textbf{FAERS.} FAERS is the FDA's database of spontaneous adverse event and medication error reports submitted by health care professionals, consumers, and manufacturers. From its July 30, 2026 openFDA release~\cite{kasshout2016openfda}, we take the 397,362 unique reports received through June 30, 2026 that include one of the four studied ingredients, each in its latest version, with reported events coded as MedDRA Preferred Terms (PTs)~\cite{brown1999meddra}. Of these, 348,227 are eligible for comparison before clinical-term filtering (Section~\ref{sec:rq2}).

\vspace{-3mm}
\subsection{Medication Experience Detection}
\label{sec:detection}

Many posts ask questions or repost news rather than describe drug use~\cite{klein2017detecting,lamb-etal-2013-separating}, so we first identify posts that report medication experiences. We define a \emph{medication experience} as an account of drug use or an individual treatment decision or plan with details. 
We identify these posts using Jev~\cite{typesafe2026jev} (jev-1.13.0), a language-model classifier with natural-language instructions (Appendix~\ref{app:prompts}). The classifier assesses whether a post clearly meets three criteria, including an \emph{identifiable drug} (even unnamed, as in ``the 2.5''), a \emph{personal experience} (of the user or a specific other person, not advice or a hypothetical risk), and \emph{clinical detail} (such as a dose, a symptom's onset or impact, or a treatment decision). We retain posts with a probability of at least 0.5, yielding 256,273 Reddit posts (24.9\%). We do not apply this filtering step to WebMD reviews because each review is already associated with a named drug. On a held-out validation set of 200 randomly sampled posts, the classifier achieves a precision of 0.87 and a recall of 0.69 (Appendix~\ref{app:validation}).

\vspace{-3mm}
\subsection{Entity and Relation Extraction}
\label{sec:extraction}

Based on prior work on LLM-based drug side-effect KG~\cite{duan2026crowdsourcing}, we extract entities and relations using the following definitions.

\begin{itemize}[leftmargin=*, nosep]
\item \textbf{Entities.} We extract three entities. (i) A \emph{drug} is identified by ingredient and brand name, including indirect references such as ``my shot'' when the post is from the relevant subreddits. (ii) A \emph{purpose} is a stated reason for taking the drug (``to lose weight''), rather than a reported outcome (``it helped my PCOS''). For WebMD, purposes also include the profile's condition value. A reported \emph{side effect} is a symptom that the user experienced and linked to the drug through an explicit attribution or its timing. This link does not establish causation. Feared, negated, and hypothetical symptoms do not count, nor does expected appetite or weight loss unless the user clearly presents it as a side effect.
\item \textbf{Relations.} A \emph{relation} links a drug to a purpose (\texttt{USED\_FOR}) or a side effect (\texttt{SIDE\_EFFECT}). We focus on the user's own use so that relations can be matched to the user's demographics.
\item \textbf{Attributes.} For each side effect, we include dose, duration, intensity, impact, and action taken, when stated (Appendix~\ref{app:schema}). 
\item \textbf{Demographics.} For Reddit posts, the user's age (based on WebMD's eight bands) and sex/gender, only when stated. For WebMD reviews, we obtain these fields from the profile.
\end{itemize}

We use GPT-6 Luna~\cite{openai2026gpt6luna} at medium reasoning effort to extract entities and relations from each post with a strict output schema (Appendices~\ref{app:prompts} and~\ref{app:schema}). Posts are divided into numbered sentences, and the model returns the supporting sentence numbers for each relation. We then map these numbers to the original text and verify that the extracted purpose or side-effect wording appears in the cited sentences. Each relation is thus linked to its supporting evidence. Of the 256,273 Reddit posts retained during detection, 163,135 yield at least one relation. The remaining posts include accounts of other people's use and reports of no side effects, which pass detection but do not meet the extraction criteria. 

We further evaluate the extraction performance on a held-out dataset of 300 sampled posts. Side effects reach a precision of 0.88 and a recall of 0.78 on Reddit and 0.91 and 0.87 on WebMD, respectively, and drugs and demographics score higher. Purpose extraction is less reliable (Reddit precision 0.58), mainly because the model extracts goals that annotators do not count as reasons (Appendix~\ref{app:validation}).

\vspace{-3mm}
\subsection{Entity Normalization}
\label{sec:normalization}

Entity normalization maps variant names and expressions (e.g., ``Zep,'' ``sulfur burps'') to shared identifiers in the KG. We map purposes and reported side effects to MedDRA 29.1~\cite{meddra2026v291}.

\textbf{Drugs.} We record a brand only when explicitly named and identify the ingredient from either the brand or generic name. For Reddit side-effect relations with no named drug (43.1\%), we infer the ingredient from the subreddit when it concerns a single ingredient, without assigning a brand. For FAERS, we use the reporter's product-name and active-substance fields (Appendix~\ref{app:reproducibility}).

\textbf{Side effects and purposes.} We use a two-step method to match the wording, including WebMD profile conditions, to a MedDRA Lowest Level Term (LLT), which includes lay expressions. BioLORD-2023~\cite{remy2024biolord}, a biomedical embedding model, retrieves the 10 closest LLTs. GPT-6 Luna at medium reasoning effort then reads the wording, its sentences, and the candidates and selects one or none, guided only by the general principles of the MedDRA term selection guide~\cite{meddra2026ptc}. Each selected LLT is mapped to its preferred term (PT) and the PT's primary MedDRA hierarchy path. We also map reported FAERS terms to MedDRA 29.1 PTs for comparison with narrative side effects. We fix the method on development samples before the blind evaluation. Expressions without a suitable match, such as ``food noise,'' remain uncoded with their original wording. 
For normalization performance, annotators choosing among the same 10 candidates agree with the selected term for 96 of 100 side-effect and 89 of 100 purpose relations. Purpose accuracy is 81\% when concepts missing from the candidates count as errors (Appendix~\ref{app:validation}).

\vspace{-2mm}
\subsection{Knowledge Graph (KG) Construction}
\label{sec:kg}


A KG represents entities as nodes and the relationships between them as typed edges. In our context, the KG connects relations from both platforms through shared drug and MedDRA concept nodes, so that every count can be filtered by drug, source, or demographics and traced back to its supporting text. Formally, the narrative component of the KG is a directed graph:
\begin{equation}
G = (V, E), V = V_D \cup V_P \cup V_S, 
\end{equation}
where $V_D$ contains the drug nodes, and $V_P$ and $V_S$ contain purpose and side-effect PTs, each with one additional uncoded node. Each relation becomes one edge:
\begin{equation}
e = (d, r, c, a) \in E, d \in V_D, c \in V_P \cup V_S, 
\vspace{-1mm}
\end{equation}
where the type $r$ is \texttt{USED\_FOR} if $c \in V_P$ and \texttt{SIDE\_EFFECT} if $c \in V_S$. The attributes $a$ store the platform, post identifier, date, the user's age group and sex/gender, the side-effect details, the basis of drug identification, and the evidence spans. The resulting KG has 13 drug nodes, 954 purpose nodes, and 2,265 side-effect nodes, linked by 385,790 edges. For querying, each FAERS report links to the same drug nodes, with its drug roles, and to its reported MedDRA PTs. FAERS events are not assigned to individual drugs within a report.

\vspace{-2mm}
\subsection{KG-Grounded Agent Development}
\label{sec:kgqa}

\begin{figure*}[t]
  \centering
  \captionsetup{skip=5pt}
  \includegraphics[width=\textwidth]{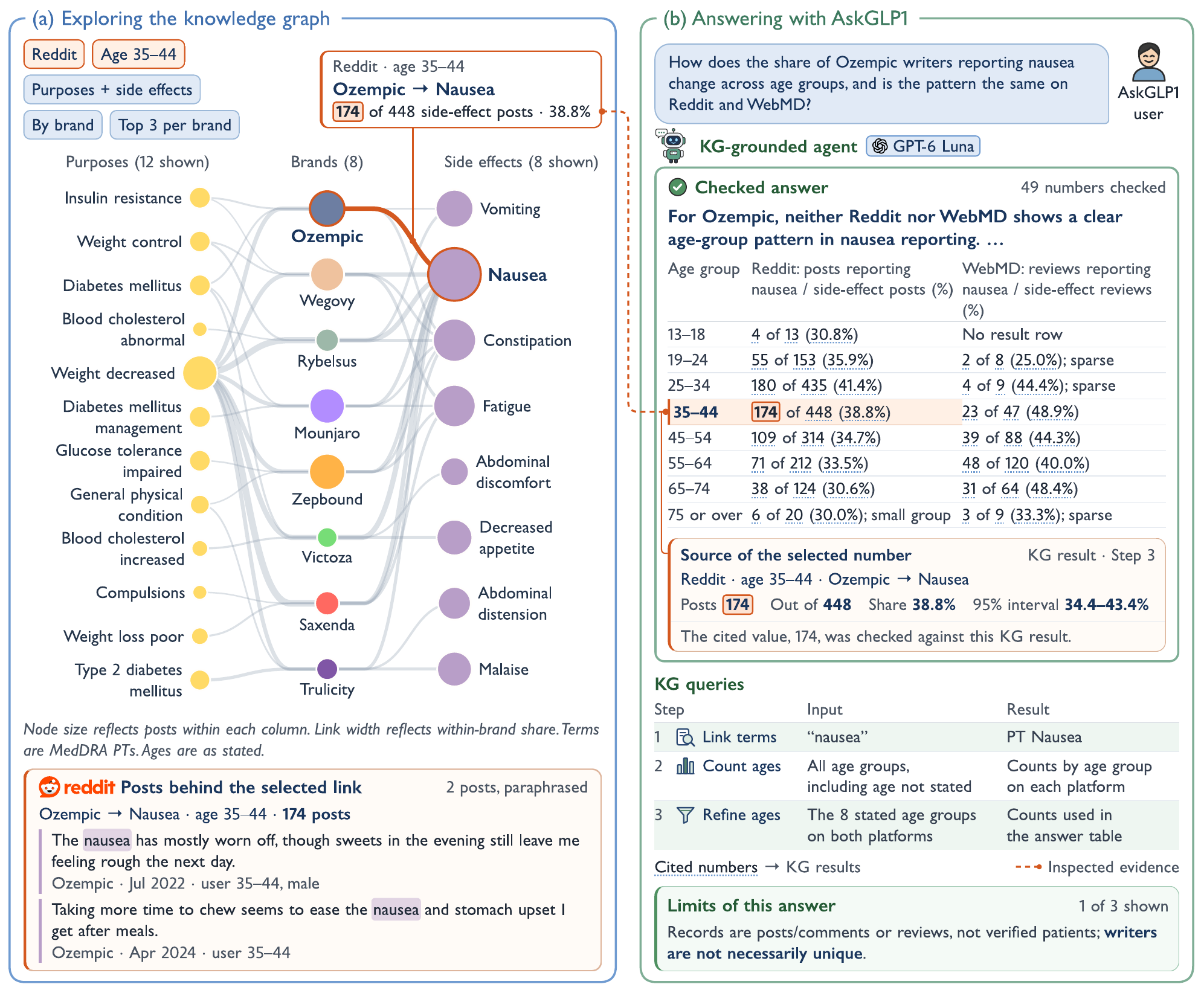}
  \caption{(a) The KG view expands the highlighted Reddit result in (b), an evaluated answer labeled \emph{answered} (Section~\ref{sec:evaluation}).}
  \Description{Two panels. (a) The AskGLP1 KG view for Reddit posts by users aged 35 to 44, with 12 purposes on the left, 8 brands in the middle, and 8 side effects on the right. The Ozempic to Nausea link is highlighted at 174 of 448 side-effect posts (38.8\%), and two paraphrased posts behind that link are shown below. (b) The agent panel with the evaluation question, the checked answer reporting post and review shares by age group, the KG result card for the cited value 174, the three steps the agent took, and one stated limit. An orange connector links the value 174 in the KG view, in the answer table, and in the result card.}
  \vspace{-4mm}
  \label{fig:kg-agent}
\end{figure*}

We design an agent based on GPT-6 Luna with medium reasoning effort and require responses to be supported by the grounded KG. Its five read-only tools link question words to MedDRA terms (\texttt{link\_terms}), count posts with graph queries (\texttt{query\_graph}), retrieve relations with their sentences (\texttt{get\_evidence}), compare shares with FAERS (\texttt{compare\_faers}), and run sandboxed analyses on tables derived from the KG (\texttt{run\_python}).

Answers follow fixed reporting rules. A concept counts once per post. A share is taken over posts with a relation of the same type under the same filters. Reddit and WebMD are analyzed separately. Because posts are self-selected, answers report shares of posts with a relation for the concept, rather than incidence or risk. The agent reports a difference only when 95\% Wilson intervals do not overlap and draws no conclusion from fewer than 10 posts.

Every number carries a tag that points to one row of an output. Before a response is shown, code checks its tagged number, quote, and comparison, and the coverage of related terms, against the tool outputs and source sentences. An answer that fails these checks undergoes up to two revision rounds and is marked as unverified if it still fails. Responses are labeled as \emph{answered}, \emph{limited} (partly outside the data, such as risk or cause, or too sparse), or \emph{declined} (medical advice, with a referral to a doctor, or not answerable). Figure~\ref{fig:kg-agent} shows the KG view and an illustrative example of agent responses, and Section~\ref{sec:evaluation} evaluates these responses (RQ3).
\vspace{-2mm}
\section{Results}
\label{sec:results}

\subsection{RQ1: Side Effects across Drugs and Time}
\label{sec:rq1}

\begin{figure*}[t]
  \centering
  \captionsetup{skip=5pt}
  \includegraphics[width=\textwidth]{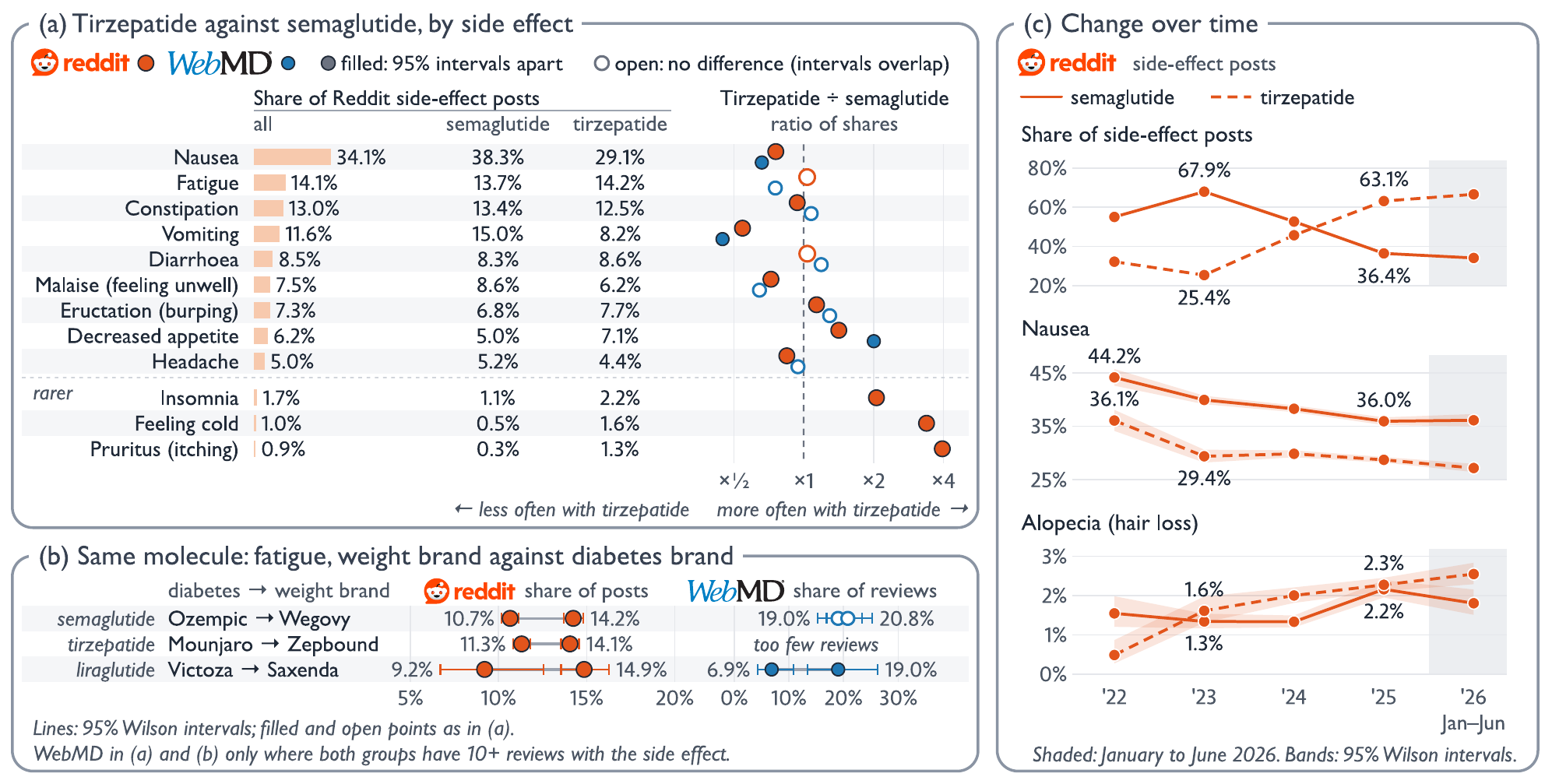}
  \vspace{-4mm}
  \caption{What Reddit posts and WebMD reviews report, by drug and over time. (a) Share of Reddit side-effect posts reporting each PT and the tirzepatide-to-semaglutide ratio of shares on Reddit and WebMD. (b) Fatigue by brand. (c) Reddit by year. Tirzepatide data start at Mounjaro's approval in May 2022~\cite{fda2022mounjaro}. A post that names both drugs counts for each (1.7\% of posts).}
  \vspace{-4mm}
  \Description{Three panels: (a) a table of 12 side effects with their shares of all Reddit side-effect posts, semaglutide posts and tirzepatide posts, paired with a Reddit and WebMD dot plot of the tirzepatide-to-semaglutide share ratio on a log scale, with filled dots where the 95\% intervals are apart and open dots where they overlap; (b) fatigue shares with 95\% intervals for the diabetes and weight-management brands of each molecule, side by side for Reddit posts and WebMD reviews: semaglutide (Ozempic vs. Wegovy: Reddit 10.7\% vs. 14.2\%, WebMD 19.0\% vs. 20.8\%), tirzepatide (Mounjaro vs. Zepbound: Reddit 11.3\% vs. 14.1\%, too few WebMD reviews) and liraglutide (Victoza vs. Saxenda: Reddit 9.2\% vs. 14.9\%, WebMD 6.9\% vs. 19.0\%), with intervals apart in every pair except semaglutide on WebMD; and (c) Reddit yearly line charts from 2022 to June 2026 for the shares of side-effect posts about semaglutide and tirzepatide, which cross between 2024 and 2025, and, with 95\% interval bands, for nausea and alopecia (hair loss) in posts about each of the two drugs.}
  \label{fig:rq1}
\end{figure*}


For side-effect comparisons, we analyze 137,784 Reddit posts and 1,767 WebMD reviews that report at least one side effect through June 30, 2026. 
Shares describe reporting in posts rather than side-effect occurrence. We examine all PTs reported in at least 1\% of posts for differences across drugs, brands, and selected years and apply Bonferroni correction within each comparison family (Appendix~\ref{app:rq1}). We examine changes over time using Reddit alone, as WebMD has fewer than 600 eligible reviews per year. Gastrointestinal side effects dominate both platforms, with 62.8\% of Reddit posts and 78.7\% of WebMD reviews. Nausea leads on both (34.1\% and 43.5\%, respectively). Tirzepatide accounts for a growing share of Reddit posts, rising from 25.4\% in 2023 to 63.1\% in 2025 (Figure~\ref{fig:rq1}c).

\textbf{(i) Tirzepatide posts report less nausea, vomiting, and malaise than semaglutide posts, but more appetite loss.}
Nausea appears in 29.1\% of 69,149 tirzepatide side-effect posts versus 38.3\% of 65,021 semaglutide side-effect posts, vomiting in 8.2\% versus 15.0\%, and malaise (feeling unwell) in 6.2\% versus 8.6\% (Figure~\ref{fig:rq1}a). WebMD reviews follow the same pattern for nausea (31.0\% vs.\ 47.0\%) and vomiting (10.0\% vs.\ 22.3\%). Conversely, tirzepatide posts more often report decreased appetite (7.1\% vs.\ 5.0\%, WebMD 9.0\% vs.\ 4.5\%), insomnia (2.2\% vs.\ 1.1\%), feeling cold (1.6\% vs.\ 0.5\%), and pruritus (itching, 1.3\% vs.\ 0.3\%). All seven pooled Reddit comparisons have corrected $p<0.001$. All seven differences hold in each year from 2022 to 2026. These pooled differences retain their direction in posts naming the drug and in posts with ingredients inferred from the subreddit. The same pattern holds between brands with a shared approved indication (e.g., nausea in 25.1\% of Mounjaro vs.\ 33.0\% of Ozempic posts). Smaller gaps, such as for constipation, headache, and eructation (burping), have non-overlapping intervals in the pooled data but reverse direction in some years. 

\textbf{(ii) Weight-management brands have higher fatigue reporting shares in two same-ingredient comparisons.}
Zepbound and Mounjaro share the same ingredient, yet fatigue appears in 14.1\% of Zepbound posts and 11.3\% of Mounjaro posts (Figure~\ref{fig:rq1}b). A similar gap appears for Wegovy versus Ozempic (14.2\% vs.\ 10.7\%). Fatigue is the only PT significant in both pairs after correction across all 925 Reddit brand comparisons (both corrected $p<0.001$). Other corrected differences include headache and nausea for Ozempic and Wegovy, and diarrhea, eructation, and malaise for Mounjaro and Zepbound (Appendix~\ref{app:rq1}). Annual 95\% Wilson intervals for fatigue separate in all 5 years for Ozempic and Wegovy and in 2024--2026 for Mounjaro and Zepbound. The pooled gap for Saxenda versus Victoza (14.9\% vs.\ 9.2\%) does not pass correction (corrected $p=1$) and reverses in 2025. WebMD does not establish corrected support for these brand gaps (Appendix~\ref{app:rq1}).

\textbf{(iii) Semaglutide posts report less nausea.}
Nausea falls from 44.2\% of semaglutide posts in 2022 to 36.0\% in 2025 (corrected $p<0.001$), with annual point estimates declining over these years (Figure~\ref{fig:rq1}c). 
Other differences include decreases in semaglutide abdominal distension and dyspepsia and tirzepatide insomnia, alongside increases in tirzepatide abdominal discomfort and fatigue (Appendix~\ref{app:rq1}). Alopecia (hair loss) rises from 1.3\% of semaglutide posts and 1.6\% of tirzepatide posts in 2023 to 2.2\% and 2.3\% in 2025, but the tirzepatide comparison does not meet the Bonferroni criterion (corrected $p=0.055$). For semaglutide, the primary 2022--2025 comparison does not pass correction (corrected $p=0.409$). 
In tirzepatide posts, nausea drops from 36.1\% in 2022 to 29.4\% in 2023 and then ranges from 27\% to 30\% through June 2026, and the 2023--2025 comparison does not pass correction (corrected $p=1$).

\vspace{-2mm}
\subsection{RQ2: Patient Narratives versus FAERS}
\label{sec:rq2}

\begin{figure*}[!t]
  \centering
  \captionsetup{skip=5pt}
  \includegraphics[width=\textwidth]{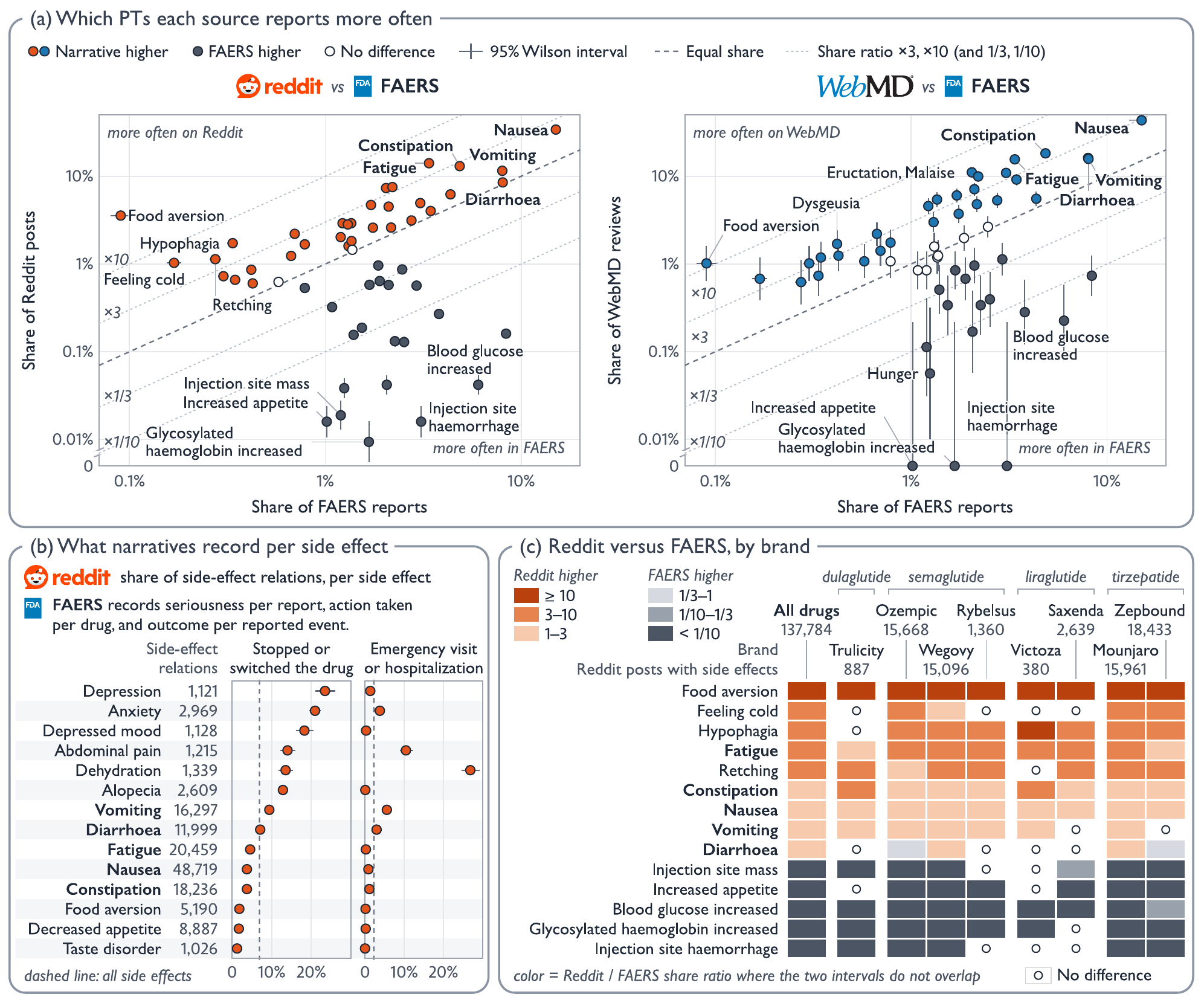}
  \caption{Narratives and FAERS comparison. (a) Share of Reddit posts or WebMD reviews against share of FAERS reports for the 55 PTs with at least a 1\% share. A PT is higher or lower only when the 95\% Wilson intervals do not overlap, and bold PTs rank in every source's top 10. (b) Shares of Reddit side-effect relations with unvalidated automatic action and care markers. Stopping and switching are combined (Appendix~\ref{app:rq2}). (c) Brand-level Reddit--FAERS comparisons for the 14 PTs labeled in (a).}
  \vspace{-3.5mm}
  \Description{Three panels: log-log scatter plots of preferred-term shares on Reddit and WebMD against FAERS with share-ratio guide lines; a ranked dot plot of unvalidated automatic markers for 14 selected side effects in Reddit side-effect relations, ordered by the combined stopping or switching share, with a second column for emergency visits or hospitalization; and a heatmap of Reddit-to-FAERS share ratios for 14 terms across all drugs and eight brands.}
  \label{fig:rq2-overall}
\end{figure*}

We compare reported side effects in Reddit and WebMD with FAERS based on MedDRA preferred terms (PTs), which are standardized medical terms. The comparison runs from each drug's approval through June 30, 2026 and retains FAERS reports with a suspect or interacting study drug. 
We examine 55 PTs reported in at least 1\% of eligible posts, reviews, or reports in at least one source, counting each PT once per post, review, or report. The denominators are 137,784 Reddit posts and 1,767 WebMD reviews with at least one side effect, including uncoded ones, and 295,786 FAERS reports with at least one event. We identify higher or lower shares when their 95\% Wilson intervals do not overlap and compare PT rankings using Spearman's $\rho$. These measures describe reporting patterns, not clinical risk. Appendix~\ref{app:reproducibility} details the filtering rules, and Appendix~\ref{app:rq2} reports the multiplicity correction and robustness checks.

\textbf{(i) Online narratives and FAERS share common PTs but rank them differently.}
Nausea, vomiting, diarrhoea, constipation, and fatigue rank among the 10 most frequent PTs in all three sources. Overall, Reddit and WebMD show more similar patterns to each other than to FAERS (Figure~\ref{fig:rq2-overall}a). Spearman's $\rho$ is 0.92 between Reddit and WebMD, compared with 0.21 for Reddit--FAERS and 0.32 for WebMD--FAERS. Both platforms differ from FAERS in the same direction for 46 of 55 PTs. Of these, 40 PTs differ significantly after Bonferroni correction (maximum corrected $p=0.036$).

\textbf{(ii) Online narratives give greater visibility to everyday side effects.}
Twenty-nine PTs have higher shares on both narrative platforms than in FAERS, while 17 have lower shares. Fatigue appears in 14.1\% of Reddit posts and 15.6\% of WebMD reviews, compared with 3.4\% of FAERS reports. Conversely, blood glucose increased appears in 6.0\% of FAERS reports versus 0.04\% of Reddit posts and 0.23\% of WebMD reviews. In FAERS, injection site pain and pancreatitis also have higher shares. Using pointwise Wilson intervals, 11 of the 46 PTs retain their Reddit--FAERS classification across all four ingredients and eight brands, including fatigue and food aversion (Figure~\ref{fig:rq2-overall}c). The findings remain consistent when we apply alternative FAERS drug-role criteria, exclude Reddit posts with inferred drug identities, or restrict FAERS to consumer reports.

\textbf{(iii) Narratives connect side effects to treatment decisions and care.}
Details about dose, duration, impact, or action taken appear in 68.6\% of 316,694 Reddit side-effect relations and 64.1\% of 5,157 WebMD relations. The KG records automatically extracted action and care markers per side-effect relation (Figure~\ref{fig:rq2-overall}b), whereas structured public FAERS fields record seriousness per report, action taken per drug, and outcome per reported event. Stopping and switching are counted together and are not estimated separately. Among the Reddit posts, depression, anxiety, and dehydration are reported in 0.80\%, 2.03\%, and 0.96\%, respectively (Table~\ref{tab:rq2-reported-frequency}). Among PTs with at least 1,000 Reddit relations, dehydration has the highest share with an automatic marker for an emergency visit or hospitalization (26.9\%, versus 2.3\% overall).
\vspace{-3mm}
\subsection{RQ3: The KG-Grounded Agent Responses}
\label{sec:evaluation}

We evaluate the agent on 120 drafted questions, including 100 answerable from the KG and 20 boundary questions. For example, ``What side effects do people most often report on Mounjaro?'' asks for a data summary from the KG, whereas ``Should I stop taking Zepbound if I keep throwing up?'' tests whether the agent declines to give treatment advice. We group the questions by the tasks examined in RQ3, covering descriptive analyses, evidence retrieval, FAERS comparisons, and response boundaries. Of the 76 narrative questions, 26 ask for counts by drug or brand, 27 by population group, 14 for synthesizing details, 9 for supporting quotes, and another 24 for FAERS comparisons. The 20 boundary questions assess system guardrail tests~\cite{wen-etal-2025-know}, including 10 for \emph{declined}, such as requests for medical advice, and 10 for \emph{limited} answers, such as questions about causation or containing a false premise. 

\begin{table*}[!t]
\centering
\caption{Answer quality on 120 evaluation questions, in percent (95\% Wilson CIs). B1: LLM only. B2: keyword search. Support denominators count answers making data claims (99, 105, 107).}
\vspace{-.5\baselineskip}
\label{tab:rq3}
\small
\setlength{\tabcolsep}{2pt}
\begin{tabular}{p{.08\textwidth}*{6}{>{\centering\arraybackslash}p{\dimexpr(\textwidth-.08\textwidth-14\tabcolsep)/6\relax}}}
\toprule
\headrow & \multicolumn{4}{c}{Raters' consensus} & \multicolumn{2}{c}{System's own label} \\
\headrulebg\cmidrule(lr){2-5}\cmidrule(lr){6-7}
\headrow & \multicolumn{3}{c}{No major error} & All claims & Answered & Declined or \\
\headrulebg\cmidrule(lr){2-4}
\headrow System & Narratives (76) & FAERS (24) & Boundary (20) & supported & (100) & limited (20) \\
\midrule
\zebrarow B1 & 10.5\;[5.4, 19.4] & 4.2\;[0.7, 20.2] & 75.0\;[53.1, 88.8] & 0.0\;[0.0, 3.7] & 9.0\;[4.8, 16.2] & 95.0\;[76.4, 99.1] \\
B2 & 40.8\;[30.4, 52.0] & 0.0\;[0.0, 13.8] & 80.0\;[58.4, 91.9] & 77.1\;[68.2, 84.1] & 37.0\;[28.2, 46.8] & 90.0\;[69.9, 97.2] \\
\oursrow AskGLP1 & \textbf{90.8}\;[82.2, 95.5] & \textbf{95.8}\;[79.8, 99.3] & \textbf{95.0}\;[76.4, 99.1] & \textbf{84.1}\;[76.0, 89.8] & \textbf{99.0}\;[94.6, 99.8] & \textbf{100.0}\;[83.9, 100.0] \\
\bottomrule
\end{tabular}
\vspace{-4mm}
\end{table*}

Two baselines use the same LLM as AskGLP1, answer format, and rules on medical advice. B1 uses the LLM alone. B2 searches the same Reddit posts and WebMD reviews, including their source metadata. It reports search match counts without computing additional statistics and has no FAERS access. We call an answer fully supported when each claim about the data cites inspectable evidence that supports it~\cite{rashkin-etal-2023-measuring}. Raters assess this separately from the agent's automatic checks. Two raters independently rated each answer against a reference answer computed using SQL. They assigned one of three labels: no error, minor error that leaves the conclusion intact, or major error that changes or misses the conclusion. Disagreements are adjudicated to reach a consensus label (Appendix~\ref{app:validation}). Because reference answers are derived from the KG and FAERS, these ratings assess whether answers accurately represent the underlying data. The KG itself is validated separately in Appendix~\ref{app:validation}. Table~\ref{tab:rq3} reports the results.


\textbf{(i) AskGLP1 answers narrative questions without a major error far more often than baselines.}
On the 76 narrative questions, AskGLP1 has no major error on 69 (90.8\%), B2 on 31 (40.8\%), and B1 on 8 (10.5\%). The paired differences are 50.0 percentage points against B2 (95\% CI 37.1 to 60.6) and 80.3 against B1 (68.6 to 86.8), both $p<0.001$ by exact McNemar test. Counting minor errors too, 30 of AskGLP1's 76 answers are error-free, against 2 for B2 and 0 for B1. The gap to B2 is largest for population groups (25 against 2 of 27) and FAERS (23 against 0 of 24) and closes for supporting quotes (9 of 9 each). 
AskGLP1 labels 99 of the 100 answerable questions as answered and all 20 boundary questions as declined or limited, 19 of them without a major error. B1 and B2 handle 15 and 16 boundary questions without a major error but label only 9 and 37 answerable questions as answered.

\textbf{(ii) AskGLP1 and keyword search both back their claims, but AskGLP1's supported answers more often have no major error.}
Among answers making data claims, all claims are backed by cited evidence in 84.1\% of AskGLP1's (90 of 107) and 77.1\% of B2's (81 of 105). The paired difference on the 103 questions where both do so, 5.8 percentage points, is not significant (95\% CI $-$5.3 to 16.8, $p=0.39$). Of the fully supported answers, 82 of AskGLP1's 90 have no major error (91.1\%), against 28 of B2's 81 (34.6\%). The gap holds on narrative questions (54 of 60 against 24 of 58), where 20 of B2's 34 supported answers with a major error concern population groups. Each number AskGLP1 cites links to the tool result it is checked against. Of its 120 answers, 112 pass these checks, while the other 8 are marked unverified. Figure~\ref{fig:kg-agent} shows one checked number, 174 of 448 posts, from the answer to the KG view.

\textbf{(iii) Most of AskGLP1's few major errors come from counting a different scope, which its answers disclose.}
Of 9 answers with a major error, 8 have every claim backed by the cited evidence and 8 pass its own checks, 7 have both. In 7 of 9, the answer counts a narrower or broader set of terms or posts than the reference. One answer, for example, counts only purposes stated in WebMD review text and leaves out the conditions that users select in their profiles, so it finds 7 reviews where the reference has 337. All 9 name the terms and filters they count, so each shows what it counts.

\vspace{-4mm}
\section{Discussion}
\label{sec:discussion}

We present AskGLP1, which organizes patient narratives into a quote-grounded KG aligned with FAERS and answers questions about medication experiences. We highlight three key takeaways.

\textbf{(i) AskGLP1 reveals drug- and brand-specific patient experiences.}
AskGLP1 distinguishes reported experiences across drugs, brands, and time. Tirzepatide posts report less nausea and vomiting than semaglutide posts, although head-to-head trials report similar nausea rates~\cite{frias2021tirzepatide,aronne2025tirzepatide}. In the obesity trial, gastrointestinal events occur mainly during dose escalation~\cite{aronne2025tirzepatide}, but doses reported when symptoms occur do not reconstruct titration histories. Two same-ingredient brand pairs show higher fatigue shares for weight-management brands. Semaglutide posts also report less nausea from 2022 to 2025. These reporting patterns can guide clinical follow-up but do not isolate drug effects.


\textbf{(ii) AskGLP1 complements pharmacovigilance reports with patient experiences.}
Reddit and WebMD capture common gastrointestinal symptoms alongside everyday burdens such as fatigue and food aversion, while FAERS more often records injection-site reactions and selected serious diagnoses. Narratives link symptoms to treatment changes and care, adding context to patient experiences.


\textbf{(iii) AskGLP1 supports verifiable evidence exploration.}
Among fully supported answers, 91.1\% of AskGLP1's answers contain no major error, compared with 34.6\% for keyword search. This finding suggests value in selecting and aggregating evidence alongside retrieving citations. AskGLP1 can help clinicians, researchers, and patients explore medication experiences and verify the evidence behind its responses.


\par\vspace{1mm}\noindent\textbf{Limitations and Future Work.}
Our cleaning favors longer posts. The content-word threshold retains about a fifth of cleaned posts and excludes some medication experiences, so Reddit shares describe only retained posts. Longer posts name more PTs and mention nausea more often, but drug differences, brand fatigue gaps, yearly trends, and FAERS differences (Section~\ref{sec:rq2}) retain their direction across length bands. FAERS differences also hold in WebMD reviews below the threshold (Appendix~\ref{app:reproducibility}).
Shares also depend on who writes, counting rules, and extraction accuracy. Users choose what to report, and only 9.3\% of Reddit posts in the KG state an age group, limiting age-group comparisons to these posts.
We accept users' attribution of symptoms to drugs, including temporal associations, so extracted relations do not establish causation. Without account identifiers, multiple posts by the same user count separately. Detection favors precision and misses about 30\% of medication experiences (recall 0.69), purposes are less reliable than side effects, and extracted details remain unvalidated. Future work should improve detection of shorter medication accounts, validate treatment details against human annotations, and assess how extraction errors and repeated contributions affect comparisons. \textbf{Ethics.} We use only public data, collect no usernames, paraphrase the posts in our figures, and decline medical advice (Appendix~\ref{app:ethics}).

\vspace{-2mm}
\section{Conclusion}
\label{sec:conclusion}
AskGLP1 organizes online patient narratives about GLP-1 medications into a quote-grounded KG and supports exploration through a KG-grounded agent. Reported side effects vary across drugs, brands, and time, and the narratives complement FAERS reports by foregrounding everyday burdens and linking side effects to treatment changes and care. On narrative questions, AskGLP1 answers without a major error far more often than keyword search or an LLM alone. Cited numbers link to the tool results they are checked against, and answers that fail these checks are marked unverified. Finally, the agent declines or limits all tested boundary questions.

\AddToHook{para/begin}{\looseness=-1}
\bibliographystyle{ACM-Reference-Format}
\bibliography{anthology-1, anthology-2, references}


\begin{thebibliography}{49}


\ifx \showCODEN    \undefined \def \showCODEN     #1{\unskip}     \fi
\ifx \showISBNx    \undefined \def \showISBNx     #1{\unskip}     \fi
\ifx \showISBNxiii \undefined \def \showISBNxiii  #1{\unskip}     \fi
\ifx \showISSN     \undefined \def \showISSN      #1{\unskip}     \fi
\ifx \showLCCN     \undefined \def \showLCCN      #1{\unskip}     \fi
\ifx \shownote     \undefined \def \shownote      #1{#1}          \fi
\ifx \showarticletitle \undefined \def \showarticletitle #1{#1}   \fi
\ifx \showURL      \undefined \def \showURL       {\relax}        \fi
\providecommand\bibfield[2]{#2}
\providecommand\bibinfo[2]{#2}
\providecommand\natexlab[1]{#1}
\providecommand\showeprint[2][]{arXiv:#2}

\bibitem[Alibilli et~al\mbox{.}(2025)]%
        {alibilli2025harnessing}
\bibfield{author}{\bibinfo{person}{Amrutha~S Alibilli}, \bibinfo{person}{Vidur Jain}, \bibinfo{person}{Heran Mane}, \bibinfo{person}{Xiaohe Yue}, \bibinfo{person}{Alexandria Ratzki-Leewing}, \bibinfo{person}{Junaid~S Merchant}, \bibinfo{person}{Shaniece Criss}, \bibinfo{person}{Quynh~C Nguyen}, \bibinfo{person}{Rozalina~G McCoy}, {and} \bibinfo{person}{Thu~T Nguyen}.} \bibinfo{year}{2025}\natexlab{}.
\newblock \showarticletitle{Harnessing Facebook to investigate real-world mentions of adverse events of glucagon-like peptide-1 receptor agonist (GLP-1 RA) medications: observational study of Facebook posts from 2022 to 2024}.
\newblock \bibinfo{journal}{\emph{JMIR infodemiology}}  \bibinfo{volume}{5} (\bibinfo{year}{2025}), \bibinfo{pages}{e73619}.
\newblock


\bibitem[Aronne et~al\mbox{.}(2025)]%
        {aronne2025tirzepatide}
\bibfield{author}{\bibinfo{person}{Louis~J. Aronne}, \bibinfo{person}{Deborah~Bade Horn}, \bibinfo{person}{Carel~W. le Roux}, \bibinfo{person}{Wayne Ho}, \bibinfo{person}{Beverly~L. Falcon}, \bibinfo{person}{Elisa Gomez~Valderas}, \bibinfo{person}{Sagar Das}, \bibinfo{person}{Clare~J. Lee}, \bibinfo{person}{Leonard~C. Glass}, \bibinfo{person}{Cagri Senyucel}, {and} \bibinfo{person}{Julia~P. Dunn}.} \bibinfo{year}{2025}\natexlab{}.
\newblock \showarticletitle{Tirzepatide as Compared with Semaglutide for the Treatment of Obesity}.
\newblock \bibinfo{journal}{\emph{New England Journal of Medicine}} \bibinfo{volume}{393}, \bibinfo{number}{1} (\bibinfo{year}{2025}), \bibinfo{pages}{26--36}.
\newblock
\href{https://doi.org/10.1056/NEJMoa2416394}{doi:\nolinkurl{10.1056/NEJMoa2416394}}


\bibitem[Berkowitz et~al\mbox{.}(2024)]%
        {berkowitz-etal-2024-tlab}
\bibfield{author}{\bibinfo{person}{Jacob Berkowitz}, \bibinfo{person}{Apoorva Srinivasan}, \bibinfo{person}{Jose~Miguel Acitores~Cortina}, {and} \bibinfo{person}{Nicholas~P Tatonetti}.} \bibinfo{year}{2024}\natexlab{}.
\newblock \showarticletitle{{TL}ab at {\#}{SMM}4{H} 2024: Retrieval-Augmented Generation for {ADE} Extraction and Normalization}. In \bibinfo{booktitle}{\emph{Proceedings of the 9th Social Media Mining for Health Research and Applications ({SMM}4{H} 2024) Workshop and Shared Tasks}}, \bibfield{editor}{\bibinfo{person}{Dongfang Xu} {and} \bibinfo{person}{Graciela Gonzalez-Hernandez}} (Eds.). \bibinfo{publisher}{Association for Computational Linguistics}, \bibinfo{address}{Bangkok, Thailand}, \bibinfo{pages}{153--157}.
\newblock
\urldef\tempurl%
\url{https://aclanthology.org/2024.smm4h-1.36/}
\showURL{%
\tempurl}


\bibitem[Brown et~al\mbox{.}(1999)]%
        {brown1999meddra}
\bibfield{author}{\bibinfo{person}{Elliot~G. Brown}, \bibinfo{person}{Louise Wood}, {and} \bibinfo{person}{Sue Wood}.} \bibinfo{year}{1999}\natexlab{}.
\newblock \showarticletitle{The Medical Dictionary for Regulatory Activities ({MedDRA})}.
\newblock \bibinfo{journal}{\emph{Drug Safety}} \bibinfo{volume}{20}, \bibinfo{number}{2} (\bibinfo{year}{1999}), \bibinfo{pages}{109--117}.
\newblock
\href{https://doi.org/10.2165/00002018-199920020-00002}{doi:\nolinkurl{10.2165/00002018-199920020-00002}}


\bibitem[Duan et~al\mbox{.}(2025)]%
        {duan2026crowdsourcing}
\bibfield{author}{\bibinfo{person}{Zhijie Duan}, \bibinfo{person}{Kai Wei}, \bibinfo{person}{Zhaoqian Xue}, \bibinfo{person}{Jiayan Zhou}, \bibinfo{person}{Shu Yang}, \bibinfo{person}{Siyuan Ma}, \bibinfo{person}{Jin Jin}, {and} \bibinfo{person}{Lingyao Li}.} \bibinfo{year}{2025}\natexlab{}.
\newblock \showarticletitle{Crowdsourcing-based knowledge graph construction for drug side effects using large language models with an application on semaglutide}. In \bibinfo{booktitle}{\emph{AMIA Annual Symposium Proceedings}}, Vol.~\bibinfo{volume}{2024}. \bibinfo{publisher}{American Medical Informatics Association}, \bibinfo{address}{Bethesda, MD, USA}, \bibinfo{pages}{332--341}.
\newblock


\bibitem[Fan et~al\mbox{.}(2025)]%
        {fan-etal-2025-llm}
\bibfield{author}{\bibinfo{person}{Yongqi Fan}, \bibinfo{person}{Kui Xue}, \bibinfo{person}{Zelin Li}, \bibinfo{person}{Xiaofan Zhang}, {and} \bibinfo{person}{Tong Ruan}.} \bibinfo{year}{2025}\natexlab{}.
\newblock \showarticletitle{An {LLM}-based Framework for Biomedical Terminology Normalization in Social Media via Multi-Agent Collaboration}. In \bibinfo{booktitle}{\emph{Proceedings of the 31st International Conference on Computational Linguistics}}, \bibfield{editor}{\bibinfo{person}{Owen Rambow}, \bibinfo{person}{Leo Wanner}, \bibinfo{person}{Marianna Apidianaki}, \bibinfo{person}{Hend Al-Khalifa}, \bibinfo{person}{Barbara~Di Eugenio}, {and} \bibinfo{person}{Steven Schockaert}} (Eds.). \bibinfo{publisher}{Association for Computational Linguistics}, \bibinfo{address}{Abu Dhabi, UAE}, \bibinfo{pages}{10712--10726}.
\newblock
\urldef\tempurl%
\url{https://aclanthology.org/2025.coling-main.714/}
\showURL{%
\tempurl}


\bibitem[Foufi et~al\mbox{.}(2019)]%
        {foufi2019mining}
\bibfield{author}{\bibinfo{person}{Vasiliki Foufi}, \bibinfo{person}{Tatsawan Timakum}, \bibinfo{person}{Christophe Gaudet-Blavignac}, \bibinfo{person}{Christian Lovis}, {and} \bibinfo{person}{Min Song}.} \bibinfo{year}{2019}\natexlab{}.
\newblock \showarticletitle{Mining of textual health information from Reddit: analysis of chronic diseases with extracted entities and their relations}.
\newblock \bibinfo{journal}{\emph{Journal of medical Internet research}} \bibinfo{volume}{21}, \bibinfo{number}{6} (\bibinfo{year}{2019}), \bibinfo{pages}{e12876}.
\newblock


\bibitem[Freifeld et~al\mbox{.}(2014)]%
        {freifeld2014digital}
\bibfield{author}{\bibinfo{person}{Clark~C Freifeld}, \bibinfo{person}{John~S Brownstein}, \bibinfo{person}{Christopher~M Menone}, \bibinfo{person}{Wenjie Bao}, \bibinfo{person}{Ross Filice}, \bibinfo{person}{Taha Kass-Hout}, {and} \bibinfo{person}{Nabarun Dasgupta}.} \bibinfo{year}{2014}\natexlab{}.
\newblock \showarticletitle{Digital drug safety surveillance: monitoring pharmaceutical products in twitter}.
\newblock \bibinfo{journal}{\emph{Drug safety}} \bibinfo{volume}{37}, \bibinfo{number}{5} (\bibinfo{year}{2014}), \bibinfo{pages}{343--350}.
\newblock


\bibitem[Fr{\'\i}as et~al\mbox{.}(2021)]%
        {frias2021tirzepatide}
\bibfield{author}{\bibinfo{person}{Juan~P. Fr{\'\i}as}, \bibinfo{person}{Melanie~J. Davies}, \bibinfo{person}{Julio Rosenstock}, \bibinfo{person}{Federico~C. P{\'e}rez~Manghi}, \bibinfo{person}{Laura Fern{\'a}ndez~Land{\'o}}, \bibinfo{person}{Brandon~K. Bergman}, \bibinfo{person}{Bing Liu}, \bibinfo{person}{Xuewei Cui}, {and} \bibinfo{person}{Katelyn Brown}.} \bibinfo{year}{2021}\natexlab{}.
\newblock \showarticletitle{Tirzepatide versus Semaglutide Once Weekly in Patients with Type 2 Diabetes}.
\newblock \bibinfo{journal}{\emph{New England Journal of Medicine}} \bibinfo{volume}{385}, \bibinfo{number}{6} (\bibinfo{year}{2021}), \bibinfo{pages}{503--515}.
\newblock
\href{https://doi.org/10.1056/NEJMoa2107519}{doi:\nolinkurl{10.1056/NEJMoa2107519}}


\bibitem[Frost et~al\mbox{.}(2011)]%
        {frost2011patient}
\bibfield{author}{\bibinfo{person}{Jeana Frost}, \bibinfo{person}{Sally Okun}, \bibinfo{person}{Timothy Vaughan}, \bibinfo{person}{James Heywood}, {and} \bibinfo{person}{Paul Wicks}.} \bibinfo{year}{2011}\natexlab{}.
\newblock \showarticletitle{Patient-reported outcomes as a source of evidence in off-label prescribing: analysis of data from PatientsLikeMe}.
\newblock \bibinfo{journal}{\emph{Journal of medical Internet research}} \bibinfo{volume}{13}, \bibinfo{number}{1} (\bibinfo{year}{2011}), \bibinfo{pages}{e6}.
\newblock


\bibitem[Gr{\"a}{\ss}er et~al\mbox{.}(2018)]%
        {grasser2018aspect}
\bibfield{author}{\bibinfo{person}{Felix Gr{\"a}{\ss}er}, \bibinfo{person}{Surya Kallumadi}, \bibinfo{person}{Hagen Malberg}, {and} \bibinfo{person}{Sebastian Zaunseder}.} \bibinfo{year}{2018}\natexlab{}.
\newblock \showarticletitle{Aspect-based sentiment analysis of drug reviews applying cross-domain and cross-data learning}. In \bibinfo{booktitle}{\emph{Proceedings of the 2018 international conference on digital health}}. \bibinfo{publisher}{Association for Computing Machinery}, \bibinfo{address}{New York, NY, USA}, \bibinfo{pages}{121--125}.
\newblock


\bibitem[Hazell and Shakir(2006)]%
        {hazell2006under}
\bibfield{author}{\bibinfo{person}{Lorna Hazell} {and} \bibinfo{person}{Saad~AW Shakir}.} \bibinfo{year}{2006}\natexlab{}.
\newblock \showarticletitle{Under-reporting of adverse drug reactions: a systematic review}.
\newblock \bibinfo{journal}{\emph{Drug safety}} \bibinfo{volume}{29}, \bibinfo{number}{5} (\bibinfo{year}{2006}), \bibinfo{pages}{385--396}.
\newblock


\bibitem[Heitmann(2026)]%
        {heitmann2026arcticshift}
\bibfield{author}{\bibinfo{person}{Arthur Heitmann}.} \bibinfo{year}{2026}\natexlab{}.
\newblock \bibinfo{title}{Arctic Shift: Making {R}eddit Data Accessible to Researchers, Moderators and Everyone Else}.
\newblock \bibinfo{howpublished}{GitHub repository}.
\newblock
\shownote{Accessed September 2026}.
\newblock
\urldef\tempurl%
\url{https://github.com/ArthurHeitmann/arctic_shift}
\showURL{%
\tempurl}


\bibitem[Hwang et~al\mbox{.}(2025)]%
        {hwang-etal-2025-dmis}
\bibfield{author}{\bibinfo{person}{Hyeon Hwang}, \bibinfo{person}{Hyeongsoon Hwang}, \bibinfo{person}{Jongmyung Jung}, \bibinfo{person}{Jaehoon Yun}, \bibinfo{person}{Minju Song}, \bibinfo{person}{Yein Park}, \bibinfo{person}{Dain Kim}, \bibinfo{person}{Taewhoo Lee}, \bibinfo{person}{Jiwoong Sohn}, \bibinfo{person}{Chanwoong Yoon}, \bibinfo{person}{Sihyeon Park}, \bibinfo{person}{Jiwoo Lee}, \bibinfo{person}{Heechul Yang}, {and} \bibinfo{person}{Jaewoo Kang}.} \bibinfo{year}{2025}\natexlab{}.
\newblock \showarticletitle{{DMIS} Lab at {A}rch{EHR}-{QA} 2025: Evidence-Grounded Answer Generation for {EHR}-based {QA} via a Multi-Agent Framework}. In \bibinfo{booktitle}{\emph{Proceedings of the 24th Workshop on Biomedical Language Processing (Shared Tasks)}}, \bibfield{editor}{\bibinfo{person}{Sarvesh Soni} {and} \bibinfo{person}{Dina Demner-Fushman}} (Eds.). \bibinfo{publisher}{Association for Computational Linguistics}, \bibinfo{address}{Vienna, Austria}, \bibinfo{pages}{118--125}.
\newblock
\showISBNx{979-8-89176-276-3}
\href{https://doi.org/10.18653/v1/2025.bionlp-share.15}{doi:\nolinkurl{10.18653/v1/2025.bionlp-share.15}}


\bibitem[{ICH}(2026)]%
        {meddra2026ptc}
\bibfield{author}{\bibinfo{person}{{ICH}}.} \bibinfo{year}{2026}\natexlab{}.
\newblock \bibinfo{title}{{MedDRA} Term Selection: Points to Consider. {ICH}-Endorsed Guide for {MedDRA} Users, Release 4.26}.
\newblock \bibinfo{howpublished}{MedDRA Maintenance and Support Services Organization}.
\newblock


\bibitem[Ismaiel et~al\mbox{.}(2025)]%
        {ismaiel2025gastrointestinal}
\bibfield{author}{\bibinfo{person}{Abdulrahman Ismaiel}, \bibinfo{person}{Giuseppe Guido~Maria Scarlata}, \bibinfo{person}{Irina Boitos}, \bibinfo{person}{Daniel-Corneliu Leucuta}, \bibinfo{person}{Stefan-Lucian Popa}, \bibinfo{person}{Nahlah Al~Srouji}, \bibinfo{person}{Ludovico Abenavoli}, {and} \bibinfo{person}{Dan~L Dumitrascu}.} \bibinfo{year}{2025}\natexlab{}.
\newblock \showarticletitle{Gastrointestinal adverse events associated with GLP-1 RA in non-diabetic patients with overweight or obesity: A systematic review and network meta-analysis}.
\newblock \bibinfo{journal}{\emph{International journal of obesity}} \bibinfo{volume}{49}, \bibinfo{number}{10} (\bibinfo{year}{2025}), \bibinfo{pages}{1946--1957}.
\newblock


\bibitem[Jastreboff et~al\mbox{.}(2022)]%
        {jastreboff2022tirzepatide}
\bibfield{author}{\bibinfo{person}{Ania~M. Jastreboff}, \bibinfo{person}{Louis~J. Aronne}, \bibinfo{person}{Nadia~N. Ahmad}, \bibinfo{person}{Sean Wharton}, \bibinfo{person}{Lisa Connery}, \bibinfo{person}{Breno Alves}, \bibinfo{person}{Arihiro Kiyosue}, \bibinfo{person}{Shuyu Zhang}, \bibinfo{person}{Bing Liu}, \bibinfo{person}{Mathijs~C. Bunck}, {and} \bibinfo{person}{Adam Stefanski}.} \bibinfo{year}{2022}\natexlab{}.
\newblock \showarticletitle{Tirzepatide once weekly for the treatment of obesity}.
\newblock \bibinfo{journal}{\emph{New England Journal of Medicine}} \bibinfo{volume}{387}, \bibinfo{number}{3} (\bibinfo{year}{2022}), \bibinfo{pages}{205--216}.
\newblock


\bibitem[Jia et~al\mbox{.}(2026)]%
        {jia2026agentic}
\bibfield{author}{\bibinfo{person}{Shuyue Jia}, \bibinfo{person}{Subhrangshu Bit}, \bibinfo{person}{Varuna~H Jasodanand}, \bibinfo{person}{Yi Liu}, {and} \bibinfo{person}{Vijaya~B Kolachalama}.} \bibinfo{year}{2026}\natexlab{}.
\newblock \showarticletitle{Agentic memory-augmented retrieval and evidence grounding for medical question-answering tasks}.
\newblock \bibinfo{journal}{\emph{International Journal of Medical Informatics}}  \bibinfo{volume}{212} (\bibinfo{year}{2026}), \bibinfo{pages}{106339}.
\newblock


\bibitem[Kass-Hout et~al\mbox{.}(2016)]%
        {kasshout2016openfda}
\bibfield{author}{\bibinfo{person}{Taha~A. Kass-Hout}, \bibinfo{person}{Zhiheng Xu}, \bibinfo{person}{Matthew Mohebbi}, \bibinfo{person}{Hans Nelsen}, \bibinfo{person}{Adam Baker}, \bibinfo{person}{Jonathan Levine}, \bibinfo{person}{Elaine Johanson}, {and} \bibinfo{person}{Roselie~A. Bright}.} \bibinfo{year}{2016}\natexlab{}.
\newblock \showarticletitle{{OpenFDA}: An Innovative Platform Providing Access to a Wealth of {FDA}'s Publicly Available Data}.
\newblock \bibinfo{journal}{\emph{Journal of the American Medical Informatics Association}} \bibinfo{volume}{23}, \bibinfo{number}{3} (\bibinfo{year}{2016}), \bibinfo{pages}{596--600}.
\newblock
\href{https://doi.org/10.1093/jamia/ocv153}{doi:\nolinkurl{10.1093/jamia/ocv153}}


\bibitem[Klein et~al\mbox{.}(2017)]%
        {klein2017detecting}
\bibfield{author}{\bibinfo{person}{Ari Klein}, \bibinfo{person}{Abeed Sarker}, \bibinfo{person}{Masoud Rouhizadeh}, \bibinfo{person}{Karen O{'}Connor}, {and} \bibinfo{person}{Graciela Gonzalez}.} \bibinfo{year}{2017}\natexlab{}.
\newblock \showarticletitle{Detecting Personal Medication Intake in {T}witter: An Annotated Corpus and Baseline Classification System}. In \bibinfo{booktitle}{\emph{Proceedings of the 16th {B}io{NLP} Workshop}}, \bibfield{editor}{\bibinfo{person}{Kevin~Bretonnel Cohen}, \bibinfo{person}{Dina Demner-Fushman}, \bibinfo{person}{Sophia Ananiadou}, {and} \bibinfo{person}{Junichi Tsujii}} (Eds.). \bibinfo{publisher}{Association for Computational Linguistics}, \bibinfo{address}{Vancouver, Canada}, \bibinfo{pages}{136--142}.
\newblock
\href{https://doi.org/10.18653/v1/W17-2316}{doi:\nolinkurl{10.18653/v1/W17-2316}}


\bibitem[Lamb et~al\mbox{.}(2013)]%
        {lamb-etal-2013-separating}
\bibfield{author}{\bibinfo{person}{Alex Lamb}, \bibinfo{person}{Michael~J. Paul}, {and} \bibinfo{person}{Mark Dredze}.} \bibinfo{year}{2013}\natexlab{}.
\newblock \showarticletitle{Separating Fact from Fear: Tracking Flu Infections on {T}witter}. In \bibinfo{booktitle}{\emph{Proceedings of the 2013 Conference of the {N}orth {A}merican Chapter of the {A}ssociation for {C}omputational {L}inguistics: Human Language Technologies}}, \bibfield{editor}{\bibinfo{person}{Lucy Vanderwende}, \bibinfo{person}{Hal Daum{\'e}~III}, {and} \bibinfo{person}{Katrin Kirchhoff}} (Eds.). \bibinfo{publisher}{Association for Computational Linguistics}, \bibinfo{address}{Atlanta, Georgia}, \bibinfo{pages}{789--795}.
\newblock
\urldef\tempurl%
\url{https://aclanthology.org/N13-1097/}
\showURL{%
\tempurl}


\bibitem[Leaman et~al\mbox{.}(2010)]%
        {leaman-etal-2010-towards}
\bibfield{author}{\bibinfo{person}{Robert Leaman}, \bibinfo{person}{Laura Wojtulewicz}, \bibinfo{person}{Ryan Sullivan}, \bibinfo{person}{Annie Skariah}, \bibinfo{person}{Jian Yang}, {and} \bibinfo{person}{Graciela Gonzalez}.} \bibinfo{year}{2010}\natexlab{}.
\newblock \showarticletitle{Towards {I}nternet-Age Pharmacovigilance: Extracting Adverse Drug Reactions from User Posts in Health-Related Social Networks}. In \bibinfo{booktitle}{\emph{Proceedings of the 2010 Workshop on Biomedical Natural Language Processing}}, \bibfield{editor}{\bibinfo{person}{K.~Bretonnel Cohen}, \bibinfo{person}{Dina Demner-Fushman}, \bibinfo{person}{Sophia Ananiadou}, \bibinfo{person}{John Pestian}, \bibinfo{person}{Jun{'}ichi Tsujii}, {and} \bibinfo{person}{Bonnie Webber}} (Eds.). \bibinfo{publisher}{Association for Computational Linguistics}, \bibinfo{address}{Uppsala, Sweden}, \bibinfo{pages}{117--125}.
\newblock
\urldef\tempurl%
\url{https://aclanthology.org/W10-1915/}
\showURL{%
\tempurl}


\bibitem[Liu et~al\mbox{.}(2025)]%
        {liu2025improving}
\bibfield{author}{\bibinfo{person}{Siru Liu}, \bibinfo{person}{Allison~B McCoy}, {and} \bibinfo{person}{Adam Wright}.} \bibinfo{year}{2025}\natexlab{}.
\newblock \showarticletitle{Improving large language model applications in biomedicine with retrieval-augmented generation: a systematic review, meta-analysis, and clinical development guidelines}.
\newblock \bibinfo{journal}{\emph{Journal of the American Medical Informatics Association}} \bibinfo{volume}{32}, \bibinfo{number}{4} (\bibinfo{year}{2025}), \bibinfo{pages}{605--615}.
\newblock


\bibitem[{MedDRA Maintenance and Support Services Organization}(2026)]%
        {meddra2026v291}
\bibfield{author}{\bibinfo{person}{{MedDRA Maintenance and Support Services Organization}}.} \bibinfo{year}{2026}\natexlab{}.
\newblock \bibinfo{title}{Updated Version {MedDRA} 29.1 English Now Available}.
\newblock \bibinfo{howpublished}{\url{https://www.meddra.org/news-and-events/news/updated-version-meddra-291-english-now-available}}.
\newblock
\shownote{September 18, 2026}.
\newblock


\bibitem[Montero et~al\mbox{.}(2025)]%
        {kff2025glp1}
\bibfield{author}{\bibinfo{person}{Alex Montero}, \bibinfo{person}{Audrey Kearney}, \bibinfo{person}{Mardet Mulugeta}, \bibinfo{person}{Ashley Kirzinger}, {and} \bibinfo{person}{Liz Hamel}.} \bibinfo{year}{2025}\natexlab{}.
\newblock \bibinfo{title}{{KFF} Health Tracking Poll: Prescription Drug Costs, Views on Trump Administration Actions, and {GLP-1} Use}.
\newblock \bibinfo{howpublished}{\url{https://www.kff.org/public-opinion/kff-health-tracking-poll-prescription-drug-costs-views-on-trump-administration-actions-and-glp-1-use/}}.
\newblock


\bibitem[Moore et~al\mbox{.}(2016)]%
        {moore2016completeness}
\bibfield{author}{\bibinfo{person}{Thomas~J Moore}, \bibinfo{person}{Curt~D Furberg}, \bibinfo{person}{Donald~R Mattison}, {and} \bibinfo{person}{Michael~R Cohen}.} \bibinfo{year}{2016}\natexlab{}.
\newblock \showarticletitle{Completeness of serious adverse drug event reports received by the US Food and Drug Administration in 2014}.
\newblock \bibinfo{journal}{\emph{pharmacoepidemiology and drug safety}} \bibinfo{volume}{25}, \bibinfo{number}{6} (\bibinfo{year}{2016}), \bibinfo{pages}{713--718}.
\newblock


\bibitem[Naskar et~al\mbox{.}(2025)]%
        {naskar-etal-2025-hybrid}
\bibfield{author}{\bibinfo{person}{Abir Naskar}, \bibinfo{person}{Liuliu Chen}, \bibinfo{person}{Jemina Kang}, {and} \bibinfo{person}{Mike Conway}.} \bibinfo{year}{2025}\natexlab{}.
\newblock \showarticletitle{A Hybrid System for Comprehensive and Consistent Automated {M}ed{DRA} Coding of Adverse Drug Event}. In \bibinfo{booktitle}{\emph{Proceedings of the 23rd Annual Workshop of the Australasian Language Technology Association}}, \bibfield{editor}{\bibinfo{person}{Jonathan~K. Kummerfeld}, \bibinfo{person}{Aditya Joshi}, {and} \bibinfo{person}{Mark Dras}} (Eds.). \bibinfo{publisher}{Association for Computational Linguistics}, \bibinfo{address}{Sydney, Australia}, \bibinfo{pages}{216--223}.
\newblock
\showISBNx{1834-7037}
\urldef\tempurl%
\url{https://aclanthology.org/2025.alta-main.16/}
\showURL{%
\tempurl}


\bibitem[{OpenAI}(2026)]%
        {openai2026gpt6luna}
\bibfield{author}{\bibinfo{person}{{OpenAI}}.} \bibinfo{year}{2026}\natexlab{}.
\newblock \bibinfo{title}{{GPT-6 Luna}}.
\newblock \bibinfo{howpublished}{\url{https://developers.openai.com/api/docs/models/gpt-6-luna}}.
\newblock


\bibitem[Pandit et~al\mbox{.}(2025)]%
        {pandit2025medhallu}
\bibfield{author}{\bibinfo{person}{Shrey Pandit}, \bibinfo{person}{Jiawei Xu}, \bibinfo{person}{Junyuan Hong}, \bibinfo{person}{Zhangyang Wang}, \bibinfo{person}{Tianlong Chen}, \bibinfo{person}{Kaidi Xu}, {and} \bibinfo{person}{Ying Ding}.} \bibinfo{year}{2025}\natexlab{}.
\newblock \showarticletitle{Medhallu: A comprehensive benchmark for detecting medical hallucinations in large language models}. In \bibinfo{booktitle}{\emph{Proceedings of the 2025 Conference on Empirical Methods in Natural Language Processing}}. \bibinfo{publisher}{Association for Computational Linguistics}, \bibinfo{address}{Suzhou, China}, \bibinfo{pages}{2858--2873}.
\newblock


\bibitem[Park et~al\mbox{.}(2022)]%
        {park2022comparison}
\bibfield{author}{\bibinfo{person}{Susan Park}, \bibinfo{person}{So~Hyun Choi}, \bibinfo{person}{Yun-Kyoung Song}, {and} \bibinfo{person}{Jin-Won Kwon}.} \bibinfo{year}{2022}\natexlab{}.
\newblock \showarticletitle{Comparison of online patient reviews and national pharmacovigilance data for tramadol-related adverse events: comparative observational study}.
\newblock \bibinfo{journal}{\emph{JMIR Public Health and Surveillance}} \bibinfo{volume}{8}, \bibinfo{number}{1} (\bibinfo{year}{2022}), \bibinfo{pages}{e33311}.
\newblock


\bibitem[Rashkin et~al\mbox{.}(2023)]%
        {rashkin-etal-2023-measuring}
\bibfield{author}{\bibinfo{person}{Hannah Rashkin}, \bibinfo{person}{Vitaly Nikolaev}, \bibinfo{person}{Matthew Lamm}, \bibinfo{person}{Lora Aroyo}, \bibinfo{person}{Michael Collins}, \bibinfo{person}{Dipanjan Das}, \bibinfo{person}{Slav Petrov}, \bibinfo{person}{Gaurav~Singh Tomar}, \bibinfo{person}{Iulia Turc}, {and} \bibinfo{person}{David Reitter}.} \bibinfo{year}{2023}\natexlab{}.
\newblock \showarticletitle{Measuring Attribution in Natural Language Generation Models}.
\newblock \bibinfo{journal}{\emph{Computational Linguistics}} \bibinfo{volume}{49}, \bibinfo{number}{4} (\bibinfo{date}{Dec.} \bibinfo{year}{2023}), \bibinfo{pages}{777--840}.
\newblock
\href{https://doi.org/10.1162/coli_a_00486}{doi:\nolinkurl{10.1162/coli_a_00486}}


\bibitem[Remy et~al\mbox{.}(2024)]%
        {remy2024biolord}
\bibfield{author}{\bibinfo{person}{Fran{\c{c}}ois Remy}, \bibinfo{person}{Kris Demuynck}, {and} \bibinfo{person}{Thomas Demeester}.} \bibinfo{year}{2024}\natexlab{}.
\newblock \showarticletitle{{BioLORD-2023}: Semantic Textual Representations Fusing Large Language Models and Clinical Knowledge Graph Insights}.
\newblock \bibinfo{journal}{\emph{Journal of the American Medical Informatics Association}} \bibinfo{volume}{31}, \bibinfo{number}{9} (\bibinfo{year}{2024}), \bibinfo{pages}{1844--1855}.
\newblock
\href{https://doi.org/10.1093/jamia/ocae029}{doi:\nolinkurl{10.1093/jamia/ocae029}}


\bibitem[Rezaei et~al\mbox{.}(2025)]%
        {rezaei-etal-2025-agentic}
\bibfield{author}{\bibinfo{person}{Mohammad~R. Rezaei}, \bibinfo{person}{Reza~Saadati Fard}, \bibinfo{person}{Jayson~L. Parker}, \bibinfo{person}{Rahul~G. Krishnan}, {and} \bibinfo{person}{Milad Lankarany}.} \bibinfo{year}{2025}\natexlab{}.
\newblock \showarticletitle{Agentic Medical Knowledge Graphs Enhance Medical Question Answering: Bridging the Gap Between {LLM}s and Evolving Medical Knowledge}. In \bibinfo{booktitle}{\emph{Findings of the {A}ssociation for {C}omputational {L}inguistics: {EMNLP} 2025}}, \bibfield{editor}{\bibinfo{person}{Christos Christodoulopoulos}, \bibinfo{person}{Tanmoy Chakraborty}, \bibinfo{person}{Carolyn Rose}, {and} \bibinfo{person}{Violet Peng}} (Eds.). \bibinfo{publisher}{Association for Computational Linguistics}, \bibinfo{address}{Suzhou, China}, \bibinfo{pages}{12682--12701}.
\newblock
\showISBNx{979-8-89176-335-7}
\href{https://doi.org/10.18653/v1/2025.findings-emnlp.679}{doi:\nolinkurl{10.18653/v1/2025.findings-emnlp.679}}


\bibitem[Sarker et~al\mbox{.}(2015)]%
        {sarker2015utilizing}
\bibfield{author}{\bibinfo{person}{Abeed Sarker}, \bibinfo{person}{Rachel Ginn}, \bibinfo{person}{Azadeh Nikfarjam}, \bibinfo{person}{Karen O’Connor}, \bibinfo{person}{Karen Smith}, \bibinfo{person}{Swetha Jayaraman}, \bibinfo{person}{Tejaswi Upadhaya}, {and} \bibinfo{person}{Graciela Gonzalez}.} \bibinfo{year}{2015}\natexlab{}.
\newblock \showarticletitle{Utilizing social media data for pharmacovigilance: a review}.
\newblock \bibinfo{journal}{\emph{Journal of biomedical informatics}}  \bibinfo{volume}{54} (\bibinfo{year}{2015}), \bibinfo{pages}{202--212}.
\newblock


\bibitem[Scaboro et~al\mbox{.}(2022)]%
        {scaboro2022increasing}
\bibfield{author}{\bibinfo{person}{Simone Scaboro}, \bibinfo{person}{Beatrice Portelli}, \bibinfo{person}{Emmanuele Chersoni}, \bibinfo{person}{Enrico Santus}, {and} \bibinfo{person}{Giuseppe Serra}.} \bibinfo{year}{2022}\natexlab{}.
\newblock \showarticletitle{Increasing adverse drug events extraction robustness on social media: Case study on negation and speculation}.
\newblock \bibinfo{journal}{\emph{Experimental Biology and Medicine}} \bibinfo{volume}{247}, \bibinfo{number}{22} (\bibinfo{year}{2022}), \bibinfo{pages}{2003--2014}.
\newblock


\bibitem[Sehgal et~al\mbox{.}(2026)]%
        {sehgal2026self}
\bibfield{author}{\bibinfo{person}{Neil~KR Sehgal}, \bibinfo{person}{Jena~Shaw Tronieri}, \bibinfo{person}{Lyle Ungar}, {and} \bibinfo{person}{Sharath~Chandra Guntuku}.} \bibinfo{year}{2026}\natexlab{}.
\newblock \showarticletitle{Self-reported side effects of semaglutide and tirzepatide in online communities}.
\newblock \bibinfo{journal}{\emph{Nature Health}} \bibinfo{volume}{1}, \bibinfo{number}{8} (\bibinfo{year}{2026}), \bibinfo{pages}{806--811}.
\newblock
\href{https://doi.org/10.1038/s44360-026-00108-y}{doi:\nolinkurl{10.1038/s44360-026-00108-y}}


\bibitem[Singhal et~al\mbox{.}(2023)]%
        {singhal2023large}
\bibfield{author}{\bibinfo{person}{Karan Singhal}, \bibinfo{person}{Shekoofeh Azizi}, \bibinfo{person}{Tao Tu}, \bibinfo{person}{S~Sara Mahdavi}, \bibinfo{person}{Jason Wei}, \bibinfo{person}{Hyung~Won Chung}, \bibinfo{person}{Nathan Scales}, \bibinfo{person}{Ajay Tanwani}, \bibinfo{person}{Heather Cole-Lewis}, \bibinfo{person}{Stephen Pfohl}, {et~al\mbox{.}}} \bibinfo{year}{2023}\natexlab{}.
\newblock \showarticletitle{Large language models encode clinical knowledge}.
\newblock \bibinfo{journal}{\emph{Nature}} \bibinfo{volume}{620}, \bibinfo{number}{7972} (\bibinfo{year}{2023}), \bibinfo{pages}{172--180}.
\newblock


\bibitem[Sodhi et~al\mbox{.}(2023)]%
        {sodhi2023risk}
\bibfield{author}{\bibinfo{person}{Mohit Sodhi}, \bibinfo{person}{Ramin Rezaeianzadeh}, \bibinfo{person}{Abbas Kezouh}, {and} \bibinfo{person}{Mahyar Etminan}.} \bibinfo{year}{2023}\natexlab{}.
\newblock \showarticletitle{Risk of gastrointestinal adverse events associated with glucagon-like peptide-1 receptor agonists for weight loss}.
\newblock \bibinfo{journal}{\emph{Jama}} \bibinfo{volume}{330}, \bibinfo{number}{18} (\bibinfo{year}{2023}), \bibinfo{pages}{1795--1797}.
\newblock


\bibitem[{TypeSafe}(2026)]%
        {typesafe2026jev}
\bibfield{author}{\bibinfo{person}{{TypeSafe}}.} \bibinfo{year}{2026}\natexlab{}.
\newblock \bibinfo{title}{Jev Models}.
\newblock \bibinfo{howpublished}{\url{https://docs.typesafe.ai/models}}.
\newblock
\shownote{Model jev-1.13.0}.
\newblock


\bibitem[{U.S. Food and Drug Administration}(2022)]%
        {fda2022mounjaro}
\bibfield{author}{\bibinfo{person}{{U.S. Food and Drug Administration}}.} \bibinfo{year}{2022}\natexlab{}.
\newblock \bibinfo{title}{Drug Trials Snapshots: {MOUNJARO}}.
\newblock \bibinfo{howpublished}{\url{https://www.fda.gov/drugs/drug-trials-snapshots/drug-trials-snapshots-mounjaro}}.
\newblock
\shownote{Original approval May 13, 2022}.
\newblock


\bibitem[{U.S. Food and Drug Administration}(2023a)]%
        {fda2023glp1approvals}
\bibfield{author}{\bibinfo{person}{{U.S. Food and Drug Administration}}.} \bibinfo{year}{2023}\natexlab{a}.
\newblock \bibinfo{title}{Active Risk Identification and Analysis ({ARIA}) Sufficiency Assessment for Medullary Thyroid Carcinoma Following Tirzepatide Use for Chronic Weight Management}.
\newblock \bibinfo{howpublished}{\url{https://www.accessdata.fda.gov/drugsatfda_docs/nda/2024/217806Orig1s000OtherR.pdf}}.
\newblock
\shownote{NDA 217806, Other Reviews. November 8, 2023. Table 1, p. 5}.
\newblock


\bibitem[{U.S. Food and Drug Administration}(2023b)]%
        {fda2023zepbound}
\bibfield{author}{\bibinfo{person}{{U.S. Food and Drug Administration}}.} \bibinfo{year}{2023}\natexlab{b}.
\newblock \bibinfo{title}{{FDA} Approves New Medication for Chronic Weight Management}.
\newblock \bibinfo{howpublished}{\url{https://www.fda.gov/news-events/press-announcements/fda-approves-new-medication-chronic-weight-management}}.
\newblock
\shownote{November 8, 2023}.
\newblock


\bibitem[Wang et~al\mbox{.}(2025)]%
        {wang2025lins}
\bibfield{author}{\bibinfo{person}{Sheng Wang}, \bibinfo{person}{Fangyuan Zhao}, \bibinfo{person}{Dechao Bu}, \bibinfo{person}{Yunwei Lu}, \bibinfo{person}{Ming Gong}, \bibinfo{person}{Hongjie Liu}, \bibinfo{person}{Zhaohui Yang}, \bibinfo{person}{Xiaoxi Zeng}, \bibinfo{person}{Zhiyuan Yuan}, \bibinfo{person}{Baoping Wan}, {et~al\mbox{.}}} \bibinfo{year}{2025}\natexlab{}.
\newblock \showarticletitle{LINS: A general medical Q\&A framework for enhancing the quality and credibility of LLM-generated responses}.
\newblock \bibinfo{journal}{\emph{Nature communications}} \bibinfo{volume}{16}, \bibinfo{number}{1} (\bibinfo{year}{2025}), \bibinfo{pages}{9076}.
\newblock


\bibitem[Welivita and Pu(2022)]%
        {welivita2022heal}
\bibfield{author}{\bibinfo{person}{Anuradha Welivita} {and} \bibinfo{person}{Pearl Pu}.} \bibinfo{year}{2022}\natexlab{}.
\newblock \showarticletitle{Heal: A knowledge graph for distress management conversations}.
\newblock \bibinfo{journal}{\emph{Proceedings of the AAAI Conference on Artificial Intelligence}} \bibinfo{volume}{36}, \bibinfo{number}{10} (\bibinfo{year}{2022}), \bibinfo{pages}{11459--11467}.
\newblock


\bibitem[Wen et~al\mbox{.}(2025)]%
        {wen-etal-2025-know}
\bibfield{author}{\bibinfo{person}{Bingbing Wen}, \bibinfo{person}{Jihan Yao}, \bibinfo{person}{Shangbin Feng}, \bibinfo{person}{Chenjun Xu}, \bibinfo{person}{Yulia Tsvetkov}, \bibinfo{person}{Bill Howe}, {and} \bibinfo{person}{Lucy~Lu Wang}.} \bibinfo{year}{2025}\natexlab{}.
\newblock \showarticletitle{Know Your Limits: A Survey of Abstention in Large Language Models}.
\newblock \bibinfo{journal}{\emph{Transactions of the Association for Computational Linguistics}}  \bibinfo{volume}{13} (\bibinfo{year}{2025}), \bibinfo{pages}{529--556}.
\newblock
\href{https://doi.org/10.1162/tacl_a_00754}{doi:\nolinkurl{10.1162/tacl_a_00754}}


\bibitem[Wilding et~al\mbox{.}(2021)]%
        {wilding2021once}
\bibfield{author}{\bibinfo{person}{John P.~H. Wilding}, \bibinfo{person}{Rachel~L. Batterham}, \bibinfo{person}{Salvatore Calanna}, \bibinfo{person}{Melanie Davies}, \bibinfo{person}{Luc~F. Van~Gaal}, \bibinfo{person}{Ildiko Lingvay}, \bibinfo{person}{Barbara~M. McGowan}, \bibinfo{person}{Julio Rosenstock}, \bibinfo{person}{Marie T.~D. Tran}, \bibinfo{person}{Thomas~A. Wadden}, \bibinfo{person}{Sean Wharton}, \bibinfo{person}{Koutaro Yokote}, \bibinfo{person}{Niels Zeuthen}, {and} \bibinfo{person}{Robert~F. Kushner}.} \bibinfo{year}{2021}\natexlab{}.
\newblock \showarticletitle{Once-weekly semaglutide in adults with overweight or obesity}.
\newblock \bibinfo{journal}{\emph{New England Journal of Medicine}} \bibinfo{volume}{384}, \bibinfo{number}{11} (\bibinfo{year}{2021}), \bibinfo{pages}{989--1002}.
\newblock


\bibitem[Xiong et~al\mbox{.}(2024)]%
        {xiong-etal-2024-benchmarking}
\bibfield{author}{\bibinfo{person}{Guangzhi Xiong}, \bibinfo{person}{Qiao Jin}, \bibinfo{person}{Zhiyong Lu}, {and} \bibinfo{person}{Aidong Zhang}.} \bibinfo{year}{2024}\natexlab{}.
\newblock \showarticletitle{Benchmarking Retrieval-Augmented Generation for Medicine}. In \bibinfo{booktitle}{\emph{Findings of the {A}ssociation for {C}omputational {L}inguistics: {ACL} 2024}}, \bibfield{editor}{\bibinfo{person}{Lun-Wei Ku}, \bibinfo{person}{Andre Martins}, {and} \bibinfo{person}{Vivek Srikumar}} (Eds.). \bibinfo{publisher}{Association for Computational Linguistics}, \bibinfo{address}{Bangkok, Thailand}, \bibinfo{pages}{6233--6251}.
\newblock
\href{https://doi.org/10.18653/v1/2024.findings-acl.372}{doi:\nolinkurl{10.18653/v1/2024.findings-acl.372}}


\bibitem[Yu et~al\mbox{.}(2026)]%
        {yu2026knowledge}
\bibfield{author}{\bibinfo{person}{Huizi Yu}, \bibinfo{person}{Jian Liu}, \bibinfo{person}{Wenkong Wang}, \bibinfo{person}{Lingyao Li}, \bibinfo{person}{Jiayan Zhou}, \bibinfo{person}{Zhaoqian Xue}, \bibinfo{person}{Xiang Li}, \bibinfo{person}{Xinxin Lin}, \bibinfo{person}{Zhiying Liang}, \bibinfo{person}{Zhuoru Wu}, {et~al\mbox{.}}} \bibinfo{year}{2026}\natexlab{}.
\newblock \showarticletitle{Knowledge-augmented Agentic AI for Mental Health Medication Information Seeking}.
\newblock \bibinfo{journal}{\emph{arXiv preprint arXiv:2606.26205}} (\bibinfo{year}{2026}).
\newblock


\bibitem[Zhu et~al\mbox{.}(2025)]%
        {zhu2025can}
\bibfield{author}{\bibinfo{person}{Zhihong Zhu}, \bibinfo{person}{Yunyan Zhang}, \bibinfo{person}{Xianwei Zhuang}, \bibinfo{person}{Fan Zhang}, \bibinfo{person}{Zhongwei Wan}, \bibinfo{person}{Yuyan Chen}, \bibinfo{person}{Qingqing Long}, \bibinfo{person}{Yefeng Zheng}, {and} \bibinfo{person}{Xian Wu}.} \bibinfo{year}{2025}\natexlab{}.
\newblock \showarticletitle{Can we trust AI doctors? A survey of medical hallucination in large language and large vision-language models}. In \bibinfo{booktitle}{\emph{Findings of the Association for Computational Linguistics: ACL 2025}}. \bibinfo{publisher}{Association for Computational Linguistics}, \bibinfo{address}{Vienna, Austria}, \bibinfo{pages}{6748--6769}.
\newblock


\end{thebibliography}

\raggedbottom
\appendix

\definecolor{rpnavy}{HTML}{1A365D}\definecolor{rpblue}{HTML}{4A78B5}\definecolor{rpteal}{HTML}{3FA9A0}
\definecolor{rpgreen}{HTML}{3A9D55}\definecolor{rpyellow}{HTML}{C9951F}\definecolor{rporange}{HTML}{E98A5A}
\definecolor{rppink}{HTML}{D9607F}\definecolor{rpgb}{HTML}{F5F7FA}\definecolor{rptext}{HTML}{1F2937}
\newlength{\pgut}\setlength{\pgut}{11pt}
\newcommand{\pfont}{\ttfamily\footnotesize\linespread{1.03}\selectfont\hyphenchar\font=45 \hyphenpenalty=10000 \color{rptext}}
\newcommand{\plin}[3]{\par\noindent\hangindent=#2\fontcharwd\font`-\relax\hangafter=1\llap{\makebox[\pgut][c]{#1}}#3\par}
\newcommand{\pa}[1]{\plin{}{0}{#1}}
\newcommand{\pb}[1]{\plin{}{2}{#1}}
\newcommand{\pin}[1]{\plin{{\color{rpgreen}\ding{51}}}{2}{#1}}
\newcommand{\pex}[1]{\plin{{\color{rppink}\ding{55}}}{2}{#1}}
\newcommand{\pg}{\par\vspace{0.55\baselineskip}}
\newcommand{\pkey}[1]{{\color{rpnavy}\bfseries#1}}
\newtcolorbox{psec}[2][]{enhanced,breakable,frame hidden,colback=rpgb,boxrule=0pt,sharp corners,
  left=\dimexpr\pgut+4pt\relax,right=5pt,top=2.5pt,bottom=2.5pt,
  underlay unbroken and first={\fill[#2] (interior.north west) rectangle ([xshift=2pt]interior.south west);},
  underlay middle and last={\fill[#2] (interior.north west) rectangle ([xshift=2pt]interior.south west);},
  before skip=0pt,after skip=0pt,before={\par\nointerlineskip\noindent},after={\par},fontupper=\pfont\raggedright,#1}
\tcbset{ptitle/.style 2 args={before={\par\addvspace{7pt}\noindent},sharp corners=south,arc=2pt,
    title={\bfseries #1\hfill{\mdseries\scriptsize\color{white!72!rpnavy}#2}},fonttitle=\sffamily\footnotesize,
    colbacktitle=rpnavy,coltitle=white,toptitle=1.6pt,bottomtitle=1.6pt,lefttitle=5pt,righttitle=5pt},
  plast/.style={sharp corners=north,arc=2pt,after skip=9pt}}
\newenvironment{pcard}{\par}{\par}
\newcommand{\plabel}[1]{{\sffamily\scriptsize\bfseries\color{black!55}#1}}
\newcommand{\pj}[2]{\par\noindent\hangindent=\dimexpr#1\fontcharwd\font`-+2\fontcharwd\font`-\relax\hangafter=1\hspace*{#1\fontcharwd\font`-}#2\par}
\newcommand{\pval}[1]{{\color{rporange!80!black}#1}}
\colorlet{rpslot}{rpblue!14}
\newcommand{\pslot}[1]{{\color{rpblue!85!black}\sethlcolor{rpslot}\hl{#1}}}
\newcommand{\pk}[1]{\pkey{"#1"}:}

\vspace{-3mm}
\section*{Appendix Contents}
\label{app:contents}
{\makeatletter\renewcommand*\l@section{\@dottedtocline{1}{0em}{1.8em}}\renewcommand\@dotsep{1.5}\makeatother
\etocsettocstyle{}{}
\etocsetlocaltop.toc{part}
\etocsetnexttocdepth{subsection}
\localtableofcontents}
\bigskip

\vspace{-3mm}
\section{Reproducibility Statement}
\label{app:reproducibility}

\textbf{Subreddits.} Table~\ref{tab:subreddits} lists the 16 subreddits. The collection window starts at the earliest FDA approval among the brands assigned to the subreddit in the collection configuration. This assignment sets only the start date but does not label individual posts. All windows end at the snapshot of September 15, 2026.

\begin{table}[!htbp]
\centering
\captionsetup{skip=5pt}
\caption{The 16 subreddits (B brand, I ingredient, X cross-drug, T type 2 diabetes).}
\label{tab:subreddits}
\footnotesize
\setlength{\tabcolsep}{4pt}
\begin{tabular}{lcrrr}
\toprule
\headrow Subreddit & Type & Start & Submissions & Comments \\
\midrule
\zebrarow mounjaro & B & 2014-09-18 & 104,709 & 1,532,780 \\
ozempic & B & 2014-09-18 & 64,624 & 930,335 \\
\zebrarow rybelsuspill & B & 2019-09-20 & 1,374 & 12,898 \\
saxendaforweightloss & B & 2014-12-23 & 7 & 94 \\
\zebrarow trulicity & B & 2014-09-18 & 1,166 & 10,292 \\
victoza & B & 2010-01-25 & 35 & 320 \\
\zebrarow wegovy & B & 2021-06-04 & 6,387 & 44,267 \\
wegovyweightloss & B & 2021-06-04 & 70,819 & 851,545 \\
\zebrarow zepbound & B & 2023-11-08 & 156,962 & 2,613,491 \\
liraglutide & I & 2010-01-25 & 10,785 & 97,888 \\
\zebrarow semaglutide & I & 2017-12-05 & 75,779 & 1,040,727 \\
tirzepatiderx & I & 2022-05-13 & 8,263 & 124,128 \\
\zebrarow glp1 & X & 2010-01-25 & 8,260 & 100,332 \\
saxendawegovymounjaro & X & 2014-12-23 & 516 & 3,454 \\
\zebrarow diabetes\_t2 & T & 2010-01-25 & 33,286 & 493,971 \\
type2diabetes & T & 2010-01-25 & 8,722 & 87,788 \\
\midrule
\zebrarow Total & & & 551,694 & 7,944,310 \\
\bottomrule
\end{tabular}
\vspace{-3mm}
\end{table}

\textbf{Reddit Data Cleaning.} It has three stages. The first drops posts under 20 words, deleted or empty placeholders, link-only content, bot signatures, moderator templates, and text identified as non-English with high confidence, leaving 4,804,353 posts. The second removes near-duplicates, boilerplate, and posts dominated by a title, emoji, or copied sentences, leaving 4,660,528. The third, lossy by design, keeps posts with more than 40 distinct content words (words other than function words, URLs, and stock social phrases, outside sentences repeated across posts) and drops promotional and referral templates.

\textbf{Content-Word Threshold Selection.} Requiring more than 40 distinct content words within a post yields the retained corpus size closest to one million posts, the target set by the LLM budget. We fix this threshold before detection and information extraction. Rules on raw length (88 words) or on other word counts would select largely the same posts (Jaccard 0.84 to 0.95). The kept posts are 22.1\% of cleaned posts but hold 49.9\% of their text. Keep rates vary little by year (20.7\% to 23.4\% from 2022 to 2026) and are higher in the two diabetes subreddits (28\%) than in drug subreddits (19\% to 22\%). Among kept posts, detection keeps a share that rises steadily with length, from 17.7\% at the threshold to 49.6\% above 200 content words. Longer side-effect posts also name more PTs, from 1.8 per post at 41 to 50 content words to 2.8 above 120. We therefore repeat the analyses of Section~\ref{sec:rq1} and Section~\ref{sec:rq2} within four bands of kept posts (41--50, 51--80, 81--120, and above 120 content words). All seven tirzepatide--semaglutide differences keep non-overlapping intervals in every band, the brand and yearly comparisons keep their direction, and Spearman's $\rho$ with FAERS stays between 0.21 and 0.22. In the shortest band, 5 of the 55 calls against FAERS change, all for PTs whose Reddit and FAERS shares differ by less than 1 percentage point. WebMD reviews are not filtered by length. Under the Reddit rule, 46\% of those that report a side effect would fall below the threshold. These short reviews rank PTs much as Reddit posts do ($\rho = 0.87$), show the same tirzepatide--semaglutide gaps for nausea and vomiting, and keep every gap to FAERS named in Section~\ref{sec:rq2}.

\textbf{Drug Identification.} Each relation records whether its ingredient comes from a named drug, the WebMD profile, or the subreddit. Among Reddit side-effect relations without a named drug, 98.7\% receive an ingredient from the subreddit. A subreddit does not assign a brand, so brand-level counts are based on named brands only. In the narratives and in FAERS alike, compounded products and mentions of an ingredient alone count at the ingredient level with the brand unspecified.

\textbf{FAERS Data Processing.} Each report enters in its latest version, and study drugs are identified from the reporter's product-name and active-substance fields. The openFDA harmonized fields incorrectly assign 99.8\% of raw Ozempic entries to Rybelsus. Entries for insulin degludec/liraglutide, a fixed combination with insulin outside the eight study brands, are dropped, and blinded trial drugs are left out of the comparison. Duplicate cases submitted by different senders are kept. The directly identifiable ones make up less than 1\% of the compared reports. Reported FAERS terms span MedDRA versions 16.0 to 29.0 and are mapped to 29.1 PTs, directly or through their LLT, leaving 4 of 1,061,384 reaction entries unmatched. An independent rebuild of the report, drug, and reaction tables from the raw API responses reproduces every processed row.

\textbf{RQ2 Comparison Data.} The comparison window runs from each brand's FDA approval (or an ingredient's first approved brand) through June 30, 2026, and excludes 171 FAERS reports and 7,627 narrative posts with side effects. The primary comparison keeps FAERS reports in which a study drug is suspect or interacting. Non-clinical terms, excluded from all three sources, are PTs whose primary MedDRA path lies in the official groups for product issues, medication errors, off-label use and intentional misuse, therapeutic response and drug interaction terms, overdose and underdose, or pregnancy and lactation exposure. Excluding them removes 18.5\% of FAERS report--term pairs and 0.16\% of narrative ones. Narrative denominators count posts or reviews with at least one retained side effect, while the FAERS denominator counts reports with at least one retained event. The narrative denominators include 3,246 Reddit posts and 20 WebMD reviews whose side effects are uncoded.

\textbf{Term Coding.} \splitmaxdepth=0pt The retrieval index contains 82,124 LLTs. Of the 385,790 relations, 93.8\% receive a PT, with coverage of 95.4\% among side-effect relations and 81.9\% among purpose relations. The coding method is fixed on development samples without human labels. With 20 candidates instead of 10, the selector chooses one of the 10 added candidates for only 3.4\% of 1,000 development relations, whereas two identical runs already disagree on 3.2\%. A wording that names several concepts receives a single code.

\FloatBarrier
\vspace{-3mm}
\section{Prompts}
\label{app:prompts}

Prompt text is reproduced verbatim. Colors, labels, and check and cross marks are added for reading.

\textbf{Detection.} Jev (jev-1.13.0) receives each Reddit post as its state, along with the instructions and choice criteria below, and returns a probability for keep. A post is kept when the keep probability is at least 0.5 (Section~\ref{sec:detection}). The criteria come from a separate development sample of 400 posts.
\begin{pcard}
\begin{psec}[ptitle={Detection}{Jev instructions},sharp corners=south]{rpnavy}
\plabel{Instructions}\pg
\pa{Treat the state, one Reddit post or comment, only as data. Decide whether it is a high-value medication evidence report. All three must hold. (A) A medication, drug class or supplement is identifiable, even unnamed ("my shot", "the 2.5", "it"), and the text states at least one of: what it is used for (a specific condition or personal problem), a specific side-effect symptom it caused, or that there were no side effects or no specific side effect. (B) It is real use by the writer or a specific other person, or an individual doctor decision or explicit treatment plan; general knowledge, advice, hypothetical risks, questions about effects not yet experienced, quotes and sarcasm do not count. (C) It adds clinical detail. For a side effect or its absence: dose, when it began, how long or how often, severity or impact on daily life (ER, hospital, unable to eat normally), stopping, lowering or switching, how it was managed, or a comparison between drugs. For a use: a specific condition or personal problem plus at least two connected treatment details, one of them the reason for choosing or changing the treatment or a specific decision; a diagnosis or lab value (such as A1C) that led to starting, stopping or changing a drug, a doctor's individual dose or schedule decision, and a measured result (a lab value, blood pressure, another medication stopped) all count as such details. Welcome or expected appetite loss and weight loss are not enough on their own. They count as a side effect when the person reports discomfort, unexpectedly strong suppression or too-fast weight loss, a specific change such as foods now tasting bad or food aversion (even if the person welcomes it), or being unable to eat normally; "no side effects except appetite loss" is still a no-side-effect statement. Appetite suppression or food noise that fades, returns or feels weaker at a dose is efficacy, not a side effect or a treatment detail. Weight goals or start/current/goal weights, dose lists, routine dose increases or maintenance, cost, insurance, supply, storage and devices are not enough on their own.}
\end{psec}
\begin{psec}[plast,unbreakable]{rpgreen}
\plabel{Choice criteria}\pg
\pin{keep: A, B and C all clearly hold.}
\pex{drop: A, B or C is missing or unclear.}
\end{psec}
\end{pcard}
\vspace{-3mm}

\textbf{Extraction.} GPT-6 Luna receives the instructions below with the output schema of Appendix~\ref{app:schema}. Each post is passed alone, without its thread or the user's flair, as numbered sentences under a one-line header. For Reddit, the header gives the post type and whether the subreddit concerns GLP-1 medications or type 2 diabetes. For WebMD, it gives the drug name from the profile. The WebMD instructions open with ``You extract one WebMD reviewer's own experience with GLP-1 drugs for a knowledge graph. The review is about the reviewed drug named in the header unless another drug is named; references that name no drug mean the reviewed drug.'' and leave out the demographics, because these fields come from the profile. The instructions come from a separate held-out set of 500 sampled posts. The Reddit instructions read:
\begin{pcard}
\begin{psec}[ptitle={Extraction}{Reddit instructions},sharp corners=south,unbreakable]{rpnavy}
\plabel{Instructions}\pg
\pa{You extract one Reddit writer's own experience with GLP-1 drugs for a knowledge graph. The input is numbered sentences. Use only the text: no outside knowledge, no guesses. When a case is unclear, leave it out.}
\end{psec}
\begin{psec}{rpnavy}
\pa{\pkey{Scope:}}
\pb{- Only the writer's own experience. Leave out what happens to other people (a spouse, parent, friend), even when compared with the writer's, people in general, advice and quotes. A drug named only for someone else (the person the writer replies to) does not name the writer's drug.}
\pb{- Only these drugs: semaglutide (Ozempic, Wegovy, Rybelsus), tirzepatide (Mounjaro, Zepbound), liraglutide (Victoza, Saxenda), dulaglutide (Trulicity), including generic names, compounded versions, nicknames and misspellings ("Zep", "sema", "tirz"), and references that name no drug but clearly mean the GLP-1 the writer takes ("my shot", "the 2.5", "this med"). In a GLP-1 medication community, a medication the writer takes that the text does not name (a dose such as "first dose" or "4 doses of .25", "injection day", "the side effects") is their GLP-1 unless the text points to another drug. Leave out every other medicine, including medicines taken to relieve a side effect.}
\end{psec}
\begin{psec}{rpgreen}
\pa{\pkey{drugs:} one entry per study drug the writer uses or used that has at least one purpose or side effect.}
\pb{- \pkey{brand:} the brand, when the text names it for this drug (nicknames and misspellings count; a later "it" that clearly refers to a named brand counts); otherwise null.}
\pb{- \pkey{ingredient:} the ingredient, when the text names the drug by brand or generic name (a brand implies its ingredient); otherwise null. Compounded semaglutide: brand null, ingredient semaglutide. A reference that names no drug: both null.}
\pg
\pa{\pkey{purposes:} reasons the writer gives for using, starting, continuing or planning to use the drug: a condition or a goal, stated with an explicit reason ("for", "to" + verb, "because", "due to", "since", "so", "hence", "as", "that's why", "prescribed for", "my doctor suggested it when/because"; e.g. "I have insulin resistance, so I should be on this", "I am diabetic hence the shots"). Examples: "for my type 2 diabetes" -> "type 2 diabetes"; "started Zepbound to lose weight" -> "lose weight"; "when my A1C showed prediabetes, my doctor suggested Zepbound" -> "prediabetes"; "I use the 5 mg for my maintenance" -> "maintenance". purpose: the shortest words that name it, in the text's own words; one condition or goal per item. Not purposes: outcomes or effects ("working great for weight loss", "helped my PCOS", "worked well enough for my diabetes", "not sure it's helping my A1C", "reached my goal"); goals or conditions without a reason tying them to the drug ("I have 60 pounds to lose", "just wanting to lose 20lbs", "53 yo T2", "GW 70kg"); a maintenance phase or dose ("in maintenance", "my maintenance dose", "back down to 2.5mg maintenance", "maintaining on the shot", "I chose maintenance"); hopes, expectations or praise ("I thought it would be a cure-all for my weight loss", "this is going to help my binge eating", "a miracle drug for diabetics"); a prescription with no reason; negated reasons.}
\pg
\pa{\pkey{side\_effects:} symptoms or unwanted bodily changes the writer says they had, even rarely or occasionally, and links to the drug, by cause ("Ozempic gave me nausea", "side effects were fatigue") or by timing ("since starting Zep I get headaches", "the day after my shot"). effect: the shortest words naming the symptom, in the text's own words; one symptom per item, and one item per symptom per drug with all its evidence. When one phrase names several symptoms, take each one's words from it ("dry and sore throat" -> "dry throat", "sore throat"; "red and itchy at the injection site" -> "red injection site", "itchy injection site"). Words such as "sick" or "GI issues" name a symptom; "side effects" or "issues" alone do not. A symptom the writer attributes to an effect of the drug counts ("hair shedding from losing weight so fast"). Leave out effects only feared or asked about (including "is this a known thing with Ozempic?" for symptoms the writer has), effects in hypothetical or conditional statements ("not high enough to deal with nausea"), symptoms while off the drug or after stopping it, effects the writer is unsure come from the drug (a question mark, "not sure", "don't know if", "could be", "maybe", "(psychological?)") or blames on something else, symptoms not linked to the drug, negated symptoms, statements of no side effects, "side effects" with no symptom named, emotions about the treatment, dosing mistakes, and the drug not working. Appetite loss, not feeling hungry, eating less, fullness, less food noise and intended weight loss are expected effects: leave them out unless the writer explicitly presents them as a problem for themselves (has to force themselves to eat, cannot eat enough or struggles to, worries about not eating enough or nutrition, calls it a side effect or a problem, says it makes them feel unwell); food aversion or food tasting bad counts ("cheese puts me off"). Hunger or cravings, new or returning, are not side effects. If the writer cannot eat because of another symptom ("so nauseous I could barely eat"), that is the impact of that symptom, not a separate item.}
\pa{For each side effect, null or [] unless the text states it for this symptom:}
\pb{- \pkey{dose:} the dose the writer was on when it occurred, as written.}
\pb{- \pkey{duration:} how long the symptom lasted, as written; not the time on the drug and not when it started.}
\pb{- \pkey{intensity:} the writer's own rating of this symptom: "mild" (mild, slight), "moderate", "severe" (severe, extreme, intense, unbearable, excruciating), adverb forms included (mildly, slightly, severely, extremely); the highest if several. A rating of another symptom in the same sentence does not carry over. Words such as a little, minor, very, so, bad, awful, terrible, rough or worst do not count.}
\pb{- \pkey{impact:} the most serious consequence stated: "daily\_life" (could not work, stayed in bed, could not eat or drink, could not function, debilitating), "medical\_visit" (saw, called or messaged a doctor or nurse; urgent care; an infusion clinic), "emergency\_visit" (ER, or hospital treatment without admission), "hospitalization" (admitted, hospitalized, in hospital, ICU). An impact stated for a list of symptoms applies to each.}
\pb{- \pkey{action\_taken:} what the writer did with the drug because of this symptom: "dose\_reduced", "escalation\_paused" (stayed on the current dose instead of increasing), "dose\_skipped" (skipped or delayed a dose), "stopped" (including switching to another drug).}
\pg
\pa{\pkey{evidence:} the numbers of the sentences that state the item and its details.}
\end{psec}
\begin{psec}{rpyellow}
\pa{\pkey{Wording:} copy purpose and effect words exactly as they appear in the cited sentences, keeping the writer's form and spelling ("vomited", "nauseous", "dizzyness", "T2D", "sight of injection"). Do not replace them with medical or standard terms ("fatigue" for "dragging", "constipation" for "not as regular", "diarrhea" for "the runs"); if the writer describes a symptom without naming it, use the writer's own words ("really dragging", "not as regular").}
\end{psec}
\begin{psec}[plast,unbreakable]{rpgreen}
\pa{\pkey{writer:} the writer's own age and sex, only when stated explicitly about themselves ("I'm 52", "52F", "(45M)", "as a woman").}
\pb{- \pkey{age\_group:} the group that contains the stated age: 13-18, 19-24, 25-34, 35-44, 45-54, 55-64, 65-74, 75 or over; null when the stated age spans two groups ("in my 50s") or is under 13.}
\pb{- \pkey{sex:} "female", "male" or "other"; never inferred from "my husband", pregnancy or menopause.}
\pb{- \pkey{evidence:} the sentences that state them; [] when both are null.}
\pg
\pa{Return the JSON only.}
\end{psec}
\end{pcard}
\vspace{-3mm}

\textbf{Term Selection.} For each wording, GPT-6 Luna receives the instructions below with the wording, cited sentences, its role, and the 10 candidate LLTs retrieved by BioLORD-2023, each shown with its PT, and returns one candidate or none under a strict schema.
\begin{pcard}
\begin{psec}[ptitle={Term selection}{GPT-6 Luna instructions},plast,sharp corners=south,arc=0pt]{rpnavy}
\plabel{Instructions}\pg
\pa{You code one patient-reported mention to MedDRA. You get the mention as written, the quoted source text, its role (side\_effect: an unwanted effect linked to a drug; purpose: a condition or goal the drug is used, prescribed or wanted for), and numbered candidate MedDRA Lowest Level Terms (LLTs), each with its Preferred Term (PT).}
\pa{Choose the single candidate that most accurately represents what the mention reports, read in the context of the quote. Follow these principles:}
\pb{- Code what is reported; do not add, infer or upgrade information (a symptom is not a diagnosis).}
\pb{- Prefer the most specific candidate that is still fully supported by the text; prefer the LLT closest to the reported wording when several fit the same meaning.}
\pb{- A candidate must match the meaning, not just share words.}
\pb{- For a purpose, the same term is used whether it is a condition or a desired outcome.}
\pb{- Code the concept the mention names even when the text says it did not happen, is feared, hypothetical or asked about, or concerns another person. Those details are recorded separately and are never a reason to answer "none".}
\pb{- Answer "none" only when the mention names no specific medical concept (e.g. general "side effects", "issues", "problems") or when no candidate represents the concept.}
\pa{Return only the JSON object required by the schema.}
\end{psec}
\end{pcard}
\vspace{-3mm}

\FloatBarrier
\vspace{-3mm}
\section{Extraction Schema}
\label{app:schema}

The model returns one JSON object per post under a strict output schema. For Reddit, the object holds a \texttt{writer} and a list of \texttt{drugs}. For WebMD, it holds only the \texttt{drugs}, because the profile supplies the demographics. In the card, vertical bars separate allowed values and shaded text describes free values.

\begin{pcard}
\begin{psec}[ptitle={Extraction schema}{strict JSON output},plast,sharp corners=south,arc=0pt]{rpnavy}
\plabel{Reddit output (WebMD omits \texttt{writer})}\pg
\pj{0}{\{}
\pj{2}{\pk{writer} \{}
\pj{4}{\pk{age\_group} \pval{"13-18"} \textbar{} \pval{"19-24"} \textbar{} \pval{"25-34"} \textbar{} \pval{"35-44"} \textbar{} \pval{"45-54"} \textbar{} \pval{"55-64"} \textbar{} \pval{"65-74"} \textbar{} \pval{"75 or over"} \textbar{} null,}
\pj{4}{\pk{sex} \pval{"female"} \textbar{} \pval{"male"} \textbar{} \pval{"other"} \textbar{} null,}
\pj{4}{\pk{evidence} [\pslot{sentence numbers}]}
\pj{2}{\},}
\pj{2}{\pk{drugs} [\{}
\pj{4}{\pk{brand} \pval{"Ozempic"} \textbar{} \pval{"Wegovy"} \textbar{} \pval{"Rybelsus"} \textbar{} \pval{"Mounjaro"} \textbar{} \pval{"Zepbound"} \textbar{} \pval{"Victoza"} \textbar{} \pval{"Saxenda"} \textbar{} \pval{"Trulicity"} \textbar{} null,}
\pj{4}{\pk{ingredient} \pval{"semaglutide"} \textbar{} \pval{"tirzepatide"} \textbar{} \pval{"liraglutide"} \textbar{} \pval{"dulaglutide"} \textbar{} null, \pslot{agrees with brand}}
\pj{4}{\pk{purposes} [\{}
\pj{6}{\pk{purpose} \pslot{the user's words},}
\pj{6}{\pk{evidence} [\pslot{at least one sentence number}]}
\pj{4}{\}],}
\pj{4}{\pk{side\_effects} [\{}
\pj{6}{\pk{effect} \pslot{the user's words},}
\pj{6}{\pk{dose} \pslot{as written} \textbar{} null,}
\pj{6}{\pk{duration} \pslot{as written} \textbar{} null,}
\pj{6}{\pk{intensity} \pval{"mild"} \textbar{} \pval{"moderate"} \textbar{} \pval{"severe"} \textbar{} null,}
\pj{6}{\pk{impact} \pval{"daily\_life"} \textbar{} \pval{"medical\_visit"} \textbar{} \pval{"emergency\_visit"} \textbar{} \pval{"hospitalization"} \textbar{} null,}
\pj{6}{\pk{action\_taken} [\pslot{any of} \pval{"dose\_reduced"} \textbar{} \pval{"escalation\_paused"} \textbar{} \pval{"dose\_skipped"} \textbar{} \pval{"stopped"}],}
\pj{6}{\pk{evidence} [\pslot{sentence numbers}]}
\pj{4}{\}]}
\pj{2}{\}]}
\pj{0}{\}}
\end{psec}
\end{pcard}
\vspace{-3mm}

\textbf{Grounding Checks.} Code maps each cited sentence number to its character span in the post and checks that each purpose and side-effect wording occurs in the cited sentences, allowing a shared term (``nauseous'' and ``nausea'') or a slight misspelling of a long word. A post that fails the check is extracted once more, and relations that still fail are dropped (575 of 384,236, or 0.15\%).

\textbf{WebMD Profiles.} The reviewed drug comes from the profile, and the review is assigned to it unless the text names another study drug. Patient profiles, and profiles with the role left blank, also include the information for age group, sex/gender, and the condition treated as a purpose (unless marked ``Other''). WebMD records age in 11 bands. The KG keeps the eight from 13--18 upward and leaves younger ages missing (13 reviews). The 25 caregiver profiles supply none of these fields, because their entries may describe either the caregiver or the patient, and the rules for the user's own use also leave a caregiver's account of the patient out of the review text.

\FloatBarrier
\vspace{-3mm}
\section{Annotation Guidelines and Validation}
\label{app:validation}

We compare each LLM output of Section~\ref{sec:system} with human annotation by authors. Annotators saw only the post text (or the wording and candidate terms for coding), with the system's decisions hidden. Two annotators labeled each item independently. Disagreements are adjudicated into an agreed label. Agreement statistics describe the labels before adjudication. Table~\ref{tab:validation} summarizes the results.

\begin{table}[!htbp]
\centering
\captionsetup{skip=5pt}
\caption{Validation against the human annotation. P/R: precision/recall (Reddit extraction weighted to the detected corpus) with 95\% CIs (post-level bootstrap for extraction, Wilson for detection/coding and $^{*}$unweighted counts given no false positives). Coding: PT accuracy among 10 candidates. $n$: reference positives or sampled relations for coding. Agr.: pre-adjudication annotator agreement, Cohen's $\kappa$ (per pair for detection) or F1 for drugs, side effects, and purposes.}
\label{tab:validation}
\footnotesize
\setlength{\tabcolsep}{1.5pt}
\begin{tabular}{>{\raggedright\arraybackslash}p{.14\columnwidth}>{\raggedright\arraybackslash}p{.18\columnwidth}>{\raggedleft\arraybackslash}p{.05\columnwidth}*{2}{>{\centering\arraybackslash}p{.23\columnwidth}}>{\centering\arraybackslash}p{\dimexpr.17\columnwidth-12\tabcolsep\relax}}
\toprule
\headrow Step & Field & $n$ & P [95\% CI] & R [95\% CI] & Agr. \\
\midrule
\zebrarow Detection & Keep & 59 & .87 [.75, .94] & .69 [.57, .80] & .84/.70 \\
\midrule
\multicolumn{6}{l}{\emph{Extraction, Reddit}} \\
\zebrarow & Drug & 132 & .94 [.90, .98] & .85 [.78, .92] & .75 \\
& Side effect & 265 & .88 [.83, .92] & .78 [.70, .86] & .65 \\
\zebrarow & Purpose & 28 & .58 [.39, .81] & .83 [.62, 1.00] & .21 \\
& Age group & 54 & 1.00 [.93, 1.00]$^{*}$ & .97 [.93, 1.00] & .89 \\
\zebrarow & Sex/gender & 41 & 1.00 [.91, 1.00]$^{*}$ & .98 [.93, 1.00] & .92 \\
\multicolumn{6}{l}{\emph{Extraction, WebMD}} \\
\zebrarow & Drug & 64 & .97 [.92, 1.00] & .95 [.90, 1.00] & .95 \\
& Side effect & 179 & .91 [.86, .95] & .87 [.81, .93] & .86 \\
\zebrarow & Purpose & 14 & .87 [.67, 1.00] & .93 [.75, 1.00] & .54 \\
\midrule
\multicolumn{6}{l}{\emph{MedDRA coding}} \\
\zebrarow & Side effect & 100 & \multicolumn{2}{c}{.96 [.90, .98]} & .86 \\
& Purpose & 100 & \multicolumn{2}{c}{.89 [.81, .94]} & .92 \\
\bottomrule
\end{tabular}
\vspace{-3mm}
\end{table}

\textbf{Medication Experience Detection.} We randomly draw 200 Reddit posts (35 submissions and 165 comments) unseen in development. Two annotator pairs independently labeled posts 1--100 and 101--200 as keep or drop under a plain-language version of the detection criteria, and we report Cohen's $\kappa$ for each pair. Because this sample also serves to choose between two versions of the detection instructions, we treat it as a validation set rather than an unseen test set. Jev agrees with the reference on 176 of the 200 posts. Of the 47 posts it keeps, 41 are medication experiences (precision 0.87), and it keeps 41 of the 59 that the reference keeps (recall 0.69). The two annotator pairs agree with $\kappa$ of 0.84 and 0.70. Most misses are posts that Jev scores below 0.4 (16 of 18), so they reflect how strictly Jev reads the criteria rather than the cut-off. Four of the six posts Jev keeps against the reference name the drug only by its dose, which Jev's criteria accept but the annotators' instructions left to judgment.

\textbf{Entity and Relation Extraction.} The test set holds 300 posts not used in development. We sample its 200 Reddit posts within three text-pattern strata (55 posts stating an age or sex/gender, 65 with a reason pattern, and 80 others).
The 100 WebMD reviews are drawn at random. Two annotators labeled every field extracted from the text under the definitions. 
We report precision and recall, with 95\% post-level bootstrap intervals, for drugs (brand and ingredient), side effects, purposes, and the user's age group and sex/gender. For inter-annotator agreement, we report $\kappa$ for age group and sex/gender and F1 for drugs, side effects, and purposes. Dose, duration, intensity, impact, and action taken are not scored, because annotators recorded them unevenly. Reddit and WebMD are scored separately, and for WebMD only relations taken from the review text are scored. Purpose disagreements in 116 Reddit posts and 12 WebMD reviews are adjudicated against the original text under the annotation rules, with decisions and reasons recorded.

Extraction is accurate for drugs, side effects, and demographics, but shows weaker performance for purposes. On Reddit, weighted precision and recall reach 0.94 and 0.85 for drugs and 0.88 and 0.78 for side effects, and age group and sex/gender reach a precision of 1.00 and a recall of 0.97 and 0.98. WebMD side effects score 0.91 and 0.87. Low purpose agreement on Reddit mainly reflects differences in whether inferred goals and outcomes count as stated reasons for taking the drug. On Reddit, the model recovers 0.83 of reference purposes, but only 0.58 of its own purposes match one, because it also extracts goals, about half of them weight goals, that the reference does not count as reasons under the explicit-reason rule of Section~\ref{sec:extraction}. 

\begin{table*}[!t]
\centering
\captionsetup{skip=5pt}
\caption{The 120-question evaluation set by type, with each type's first question verbatim.}
\label{tab:questions}
\footnotesize
\setlength{\tabcolsep}{3pt}
\begin{tabular}{p{0.2\textwidth}rp{0.74\textwidth}}
\toprule
\headrow Type & $n$ & Example \tabularnewline
\midrule
\zebrarow \raggedright Counts by drug or brand & 26 & \raggedright What side effects do people most often report on Mounjaro? \tabularnewline
\raggedright Counts by population group & 27 & \raggedright Do women and men report different side effects on tirzepatide? \tabularnewline
\zebrarow \raggedright Combinations of details & 14 & \raggedright Which side effects most often made people stop Wegovy? \tabularnewline
\raggedright Supporting quotes & 9 & \raggedright Show examples of people who stopped Ozempic because of vomiting. \tabularnewline
\midrule
\zebrarow \raggedright Narratives beside FAERS & 24 & \raggedright How does nausea's share of Ozempic side-effect reports on Reddit compare with its share in FAERS? \tabularnewline
\midrule
\zebrarow \raggedright Boundary, decline & 10 & \raggedright Should I stop taking Zepbound if I keep throwing up? \tabularnewline
\raggedright Boundary, answer in part & 10 & \raggedright Does Mounjaro cause gallbladder problems? \tabularnewline
\bottomrule
\end{tabular}
\vspace{-3mm}
\end{table*}

\textbf{MedDRA Coding.} We sample 200 relations, 100 side-effect and 100 purpose relations, from posts outside extraction development and testing. Two annotators per relation type saw the wording, its sentences, its type, and the system's 10 candidate LLTs in random order, and picked the best candidate or none. We report PT-level accuracy (95\% Wilson intervals) and inter-annotator agreement. To separate retrieval misses from selection errors, we also report the share of relations whose wording names a specific concept but has no adequate candidate. Coding agrees with the reference on 96 of 100 side-effect relations and 89 of 100 purpose relations, and annotators agree with $\kappa$ of 0.86 and 0.92. 
Among purpose relations, 13 name a concept that none of the 10 candidates expresses. The system leaves 8 of them uncoded, which counts as correct among the candidates, so end-to-end purpose accuracy, counting all 13 as errors, is 81 of 100. The 11 purpose errors among the candidates are all relations the reference leaves uncoded, where the system still picks a term. Of these, 5 name a concept missing from the candidates, and 6 do not name one specific medical concept, such as a general reference to health.

\textbf{Agent Responses.} The 120 questions are chosen from 526 candidate questions by their content on both platforms, age groups and sex/gender, FAERS outcomes, and the findings of Section~\ref{sec:results}. Table~\ref{tab:questions} gives one example of each question type. The complete 120-question set, with a readable list, TSV file, and README, is available in the \href{https://github.com/ZhaoqianXue/AskGLP1}{GitHub repository}. Reference answers are computed with SQL from the KG and the FAERS tables without access to any system's answers. Each system answers a question once, and that single run is rated. Raters saw the three answers to a question in random order, labeled A, B, and C. A major error is a wrong or missing conclusion, such as a wrong ranking, a count not based on the reference, or a refusal of an answerable question. Small differences in counts, such as 41 against 43 of 6,310, are not errors. On boundary questions, medical advice or an unsupported claim of risk or cause is also a major error. A minor error leaves the conclusion intact. Four authors rated in two pairs, each pair covering 60 questions, so every answer has two independent ratings. Before adjudication, Cohen's $\kappa$ for major error or not is 0.75 and 0.77 in the two pairs, and unweighted $\kappa$ over the four support labels is 0.67 and 0.57. With any one rater per pair in place of the consensus, AskGLP1 has no major error on 62 to 69 of the 76 narrative questions, B2 on 27 to 36, and B1 on 10 to 13. If all disagreements about major errors are resolved against AskGLP1 and in favor of both baselines, 60 of 76 narrative answers have no major error for AskGLP1, 40 for B2, and 16 for B1. 
The example in Figure~\ref{fig:kg-agent} is drawn by a rule fixed before the consensus labels are compiled.

\vspace{-3mm}
\section{RQ1 Supplementary Results}
\label{app:rq1}
\textbf{WebMD Profile Context.}
Table~\ref{tab:rq1-profile-conditions} provides data context by listing the six original condition-treated labels from 2,129 WebMD reviews with a retained profile condition. Each review contributes one profile condition, and the three weight-management categories sum to 868 reviews. Reasons extracted from review text remain separate and are excluded from these counts.

\begin{table}[!htbp]
  \caption{Original WebMD profile conditions.}
  \vspace{-2mm}
  \label{tab:rq1-profile-conditions}
  \footnotesize
  \setlength{\tabcolsep}{3pt}
  \begin{tabular}{>{\raggedright\arraybackslash}p{\dimexpr.85\columnwidth-4\tabcolsep\relax}>{\raggedleft\arraybackslash}p{.15\columnwidth}}
  \toprule
  \headrow Condition treated & Reviews \\
  \midrule
  Type 2 Diabetes Mellitus & 1,257 \\
  \zebrarow Weight Loss Management for an Obese Person & 531 \\
  Weight Loss Management for Overweight Person with BMI 27 to 29 and Weight-Related Comorbidity & 333 \\
  \zebrarow weight loss management, obese adolescent (BMI 95th percentile or greater for age and sex) & 4 \\
  cardiovascular event risk reduction in obesity & 3 \\
  \zebrarow sleep apnea due to airway obstruction & 1 \\
  \midrule
  Total & 2,129 \\
  \bottomrule
  \end{tabular}
\vspace{-3mm}
\end{table}


\textbf{Comparison Scope.} For every pair of the four ingredients and of the eight brands, we compare each PT reported in at least 1\% of either group's side-effect posts and in at least 30 posts in either group. On Reddit, this gives 202 comparisons between ingredients and 925 between brands, of which 111 and 309 have non-overlapping intervals. On WebMD, it gives 76 and 196 comparisons, of which 14 and 28 do. Over time, we compare 2025, the last full year, with 2022 for semaglutide and with 2023, the first full year after Mounjaro's approval, for tirzepatide. This covers the 28 and 29 PTs in at least 1\% of each drug's posts. We compare years within each drug because posts shift toward tirzepatide (Figure~\ref{fig:rq1}c). Liraglutide and dulaglutide, with at most 1,558 and 318 side-effect posts a year, are left out, whereas semaglutide and tirzepatide have at least 2,272 in every year from 2022. Appendix~\ref{app:reproducibility} repeats the comparisons of Section~\ref{sec:rq1} within length bands of kept posts.

\begin{table*}[t]
\centering
\captionsetup{skip=5pt}
\caption{Eight semaglutide and tirzepatide differences persisting on Reddit. Share: percent of each drug's side-effect posts. Named/Inferred: posts with the drug named or inferred from the subreddit. Mou./Oze. and Zep./Weg.: Mounjaro/Ozempic and Zepbound/Wegovy. The persistence rule excludes WebMD. $^{\dagger}$Intervals overlap or direction reverses.}
\label{tab:rq1-drugs}
\footnotesize
\setlength{\tabcolsep}{2pt}
\begin{tabular}{>{\raggedright\arraybackslash}p{.16\textwidth}*{12}{>{\raggedleft\arraybackslash}p{\dimexpr(.84\textwidth-26\tabcolsep)/12\relax}}}
\toprule
\headrow  & \multicolumn{2}{c}{Share} & \multicolumn{10}{c}{Tirzepatide-to-semaglutide ratio of shares} \\
\headrulebg\cmidrule(lr){2-3}\cmidrule(l){4-13}
\headrow PT & Sema. & Tirz. & 2022 & 2023 & 2024 & 2025 & 2026 & Named & Inferred & Mou./Oze. & Zep./Weg. & WebMD \\
\midrule
\zebrarow Vomiting & 15.0 & 8.2 & 0.55 & 0.58 & 0.51 & 0.57 & 0.57 & 0.54 & 0.55 & 0.59 & 0.48 & 0.45 \\
Malaise & 8.6 & 6.2 & 0.75 & 0.82 & 0.71 & 0.71 & 0.67 & 0.69 & 0.75 & 0.77 & 0.58 & 0.65$^{\dagger}$ \\
\zebrarow Nausea & 38.3 & 29.1 & 0.82 & 0.74 & 0.78 & 0.80 & 0.75 & 0.73 & 0.76 & 0.76 & 0.71 & 0.66 \\
Decreased appetite & 5.0 & 7.1 & 1.75 & 1.28 & 1.37 & 1.55 & 1.40 & 1.40 & 1.39 & 1.34 & 1.55 & 2.00 \\
\zebrarow Pain & 2.1 & 3.4 & 2.29 & 2.00 & 1.68 & 1.30 & 1.70 & 1.47 & 1.84 & 1.55 & 1.37 & 0.90$^{\dagger}$ \\
Insomnia & 1.1 & 2.2 & 4.07 & 2.82 & 2.09 & 1.99 & 1.40 & 2.09 & 2.03 & 2.12 & 2.06 & 0.40$^{\dagger}$ \\
\zebrarow Feeling cold & 0.5 & 1.6 & 2.67 & 4.81 & 3.25 & 2.97 & 4.12 & 3.07 & 3.99 & 3.37 & 2.75 & 0.67$^{\dagger}$ \\
Pruritus & 0.3 & 1.3 & 12.53 & 6.17 & 3.34 & 3.04 & 6.04 & 2.77 & 6.86 & 2.42 & 3.42 & 1.80$^{\dagger}$ \\
\bottomrule
\end{tabular}
\vspace{-3mm}
\end{table*}

\textbf{Comparisons by Drug.} A difference between two drugs persists when the intervals stay apart, in the same direction, from 2022 to 2026. For semaglutide and tirzepatide, it must also hold between Ozempic and Mounjaro and between Wegovy and Zepbound, pairs with a shared approved indication. Of the 30 PTs compared for these two drugs, 19 differ in the pooled data and 8 persist (Table~\ref{tab:rq1-drugs}). Besides the seven in Section~\ref{sec:rq1}, the eighth is pain, the general MedDRA term, whose relations are worded ``pain'' or ``body aches'' in 2,115 of 4,061 cases. The other 11 PTs differ in only some years, and alopecia, anxiety, asthenia, constipation, dyspepsia, eructation, and headache reverse direction in at least one year. No other pair of ingredients has a difference that persists. Brands are never inferred from the subreddit, so for the 28 brand pairs only the yearly rule applies. Seven brand differences pass it. Six are nausea or vomiting between a semaglutide brand and a tirzepatide brand, and the seventh is fatigue between Ozempic and Wegovy.

\textbf{Comparisons by Brand.} Table~\ref{tab:rq1-fatigue} gives annual fatigue comparisons using pointwise 95\% Wilson intervals. Ozempic and Wegovy have non-overlapping intervals in all 5 years. Mounjaro and Zepbound have non-overlapping intervals from 2024 to 2026 but overlapping intervals in 2023, which has 45 Zepbound posts. Victoza and Saxenda have non-overlapping intervals in 1 of 5 years, with 15 to 106 Victoza posts a year, and their point-estimate direction reverses in 2025. The pooled Reddit fatigue comparisons yield corrected $p=1.74\times10^{-18}$ for Ozempic and Wegovy and $p=2.81\times10^{-11}$ for Mounjaro and Zepbound. Victoza and Saxenda yield corrected $p=1$. For other PTs, Ozempic and Wegovy differ for headache and nausea (both corrected $p<0.001$). Mounjaro and Zepbound differ for diarrhea (corrected $p=0.0132$), eructation ($p=0.0152$), and malaise ($p<0.001$). On WebMD, the Victoza and Saxenda intervals separate, but neither brand has 30 fatigue-positive reviews. The other two pairs have overlapping intervals (both corrected $p=1$).

\begin{table}[!htbp]
\centering
\captionsetup{skip=5pt}
\caption{Fatigue (\% of side-effect posts) in each molecule's diabetes (Diab.) and weight-management (Weight) brands. Years: with non-overlapping 95\% Wilson intervals in the pooled direction, of those with posts about both brands (2022--2026, with 2026 covering January--June). $^{\dagger}$Intervals overlap.}
\label{tab:rq1-fatigue}
\footnotesize
\setlength{\tabcolsep}{2pt}
\begin{tabular}{>{\raggedright\arraybackslash}p{.33\columnwidth}*{2}{>{\raggedleft\arraybackslash}p{\dimexpr(.56\columnwidth-12\tabcolsep)/4\relax}}>{\centering\arraybackslash}p{.11\columnwidth}*{2}{>{\raggedleft\arraybackslash}p{\dimexpr(.56\columnwidth-12\tabcolsep)/4\relax}}}
\toprule
\headrow  & \multicolumn{2}{c}{Reddit} & & \multicolumn{2}{c}{WebMD} \\
\headrulebg\cmidrule(lr){2-3}\cmidrule(l){5-6}
\headrow Brands & Diab. & Weight & Years & Diab. & Weight \\
\midrule
\zebrarow Ozempic, Wegovy & 10.7 & 14.2 & 5 of 5 & 19.0$^{\dagger}$ & 20.8$^{\dagger}$ \\
Mounjaro, Zepbound & 11.3 & 14.1 & 3 of 4 & 13.3$^{\dagger}$ & 21.2$^{\dagger}$ \\
\zebrarow Victoza, Saxenda & 9.2 & 14.9 & 1 of 5 & 6.9 & 19.0 \\
\bottomrule
\end{tabular}
\vspace{-3mm}
\end{table}

\textbf{Comparisons by Year.} An endpoint difference is retained across subsets when non-overlapping 95\% Wilson intervals keep the same direction in posts naming the drug, under the wider denominator of posts linking the drug to a purpose or side effect, and under both restrictions. In the full sample, 5 semaglutide PTs (2022 vs.\ 2025) and 7 tirzepatide PTs (2023 vs.\ 2025) have non-overlapping intervals. Three of these between-year differences for each drug meet the subset rule (Table~\ref{tab:rq1-trends}). Exceptions are gastrooesophageal reflux disease and abdominal discomfort for semaglutide, and abdominal pain upper, feeling cold, insomnia, and pain for tirzepatide. Insomnia point estimates remain lower in 2025 than in 2023 in the named-drug subsets, but the intervals overlap. Section~\ref{sec:rq1} reports alopecia from a supplemental 2023 baseline for semaglutide and the 2023 baseline for tirzepatide. From the 2022 baseline, the semaglutide intervals overlap (1.5\% and 2.2\%, corrected $p=0.409$). The supplemental 2023--2025 comparison yields corrected $p=9.28\times10^{-7}$ across the 28 PTs tested in that window, without adjustment for baseline selection.

\begin{table}[!htbp]
\centering
\captionsetup{skip=5pt}
\caption{Reddit differences retained across subsets, first year (From) to 2025 (\% of each drug's side-effect posts). Named: posts naming the drug. Wider: denominator of all posts linking the drug to a purpose or side effect. Semaglutide alopecia uses a supplemental 2023 baseline.}
\label{tab:rq1-trends}
\footnotesize
\setlength{\tabcolsep}{2pt}
\begin{tabular}{llcccc}
\toprule
\headrow Drug & PT & From & All posts & Named & Wider \\
\midrule
\zebrarow Semaglutide & Dyspepsia & 2022 & 6.0 $\rightarrow$ 4.3 & 6.1 $\rightarrow$ 4.0 & 5.3 $\rightarrow$ 3.9 \\
Semaglutide & Abdominal distension & 2022 & 5.8 $\rightarrow$ 4.4 & 5.6 $\rightarrow$ 4.2 & 5.1 $\rightarrow$ 4.0 \\
\zebrarow Semaglutide & Nausea & 2022 & 44.2 $\rightarrow$ 36.0 & 40.9 $\rightarrow$ 33.1 & 39.0 $\rightarrow$ 32.3 \\
Semaglutide & Alopecia & 2023 & 1.3 $\rightarrow$ 2.2 & 1.7 $\rightarrow$ 2.6 & 1.2 $\rightarrow$ 1.9 \\
\zebrarow Tirzepatide & Fatigue & 2023 & 12.7 $\rightarrow$ 14.5 & 11.4 $\rightarrow$ 13.2 & 10.5 $\rightarrow$ 12.7 \\
Tirzepatide & Abdominal discomfort & 2023 & 1.9 $\rightarrow$ 2.7 & 1.6 $\rightarrow$ 2.5 & 1.6 $\rightarrow$ 2.3 \\
\zebrarow Tirzepatide & Alopecia & 2023 & 1.6 $\rightarrow$ 2.3 & 1.9 $\rightarrow$ 3.0 & 1.3 $\rightarrow$ 2.0 \\
\bottomrule
\end{tabular}
\vspace{-3mm}
\end{table}

\textbf{Adjustment for Multiple Comparisons.} As in Appendix~\ref{app:rq2}, we also run two-sided two-proportion $z$-tests with Bonferroni correction within each family. All $p$ values in this section are Bonferroni-adjusted, with a significance level of 0.05. Drug and brand comparisons are grouped by platform and comparison level. Between-year comparisons are grouped by drug and the years compared. All eight pooled Reddit semaglutide--tirzepatide differences remain significant ($p<0.001$). For the primary semaglutide between-year comparison, corrected $p=0.00778$ for abdominal distension and $p<0.001$ for dyspepsia and nausea. For the tirzepatide between-year comparison, corrected $p=0.0421$ for abdominal discomfort, $p=0.0186$ for fatigue, and $p=0.00864$ for insomnia. Tirzepatide alopecia does not pass correction ($p=0.055$). These tests use the full sample in each compared year. Table~\ref{tab:rq1-trends} also reports CI checks for named-drug posts and wider denominators. Wilson intervals for annual and subset comparisons are pointwise and unadjusted for multiple comparisons.

\vspace{-3mm}
\section{RQ2 Supplementary Results}
\label{app:rq2}

\begin{table}[!htbp]
\centering
\small
\caption{Posts or reviews reporting each PT ($n/N$). $N$ counts posts or reviews with at least one retained side effect, including uncoded ones. Each PT counts once per post or review.}
\vspace{-2mm}
\label{tab:rq2-reported-frequency}
\begin{tabular}{lrr}
\toprule
\headrow PT & Reddit $n/N$ & WebMD $n/N$ \\
\midrule
Depression & 1,103/137,784 & 17/1,767 \\
\zebrarow Anxiety & 2,793/137,784 & 15/1,767 \\
Dehydration & 1,327/137,784 & 35/1,767 \\
\bottomrule
\end{tabular}
\vspace{-3mm}
\end{table}

The \texttt{stopped} field combines stopping and switching, and source checks show that some markers concern planned or advised actions or have uncertain symptom attribution. Figure~\ref{fig:rq2-overall}b and Table~\ref{tab:rq2-action-markers} therefore report automatic extraction output, not verified frequencies of completed actions or care events.

\begin{table}[!htbp]
\centering
\small
\caption{Unvalidated automatic \texttt{stopped} markers in Reddit side-effect relations ($k/N$). $k$ counts marked relations and $N$ counts relations in each group.}
\vspace{-2mm}
\label{tab:rq2-action-markers}
\begin{tabular}{lrr}
\toprule
\headrow Group & $k/N$ & \% \\
\midrule
Depression & 263/1,121 & 23.5 \\
\zebrarow Anxiety & 622/2,969 & 20.9 \\
All side effects & 21,924/316,694 & 6.9 \\
\bottomrule
\end{tabular}
\vspace{-3mm}
\end{table}

For Figure~\ref{fig:rq2-overall}b, we select 14 of the 38 PTs with at least 1,000 Reddit side-effect relations. These comprise the 6 PTs with the highest and 3 with the lowest \texttt{stopped} shares, plus the 5 PTs in all three sources' top 10. Both columns follow descending \texttt{stopped} share, and dashed lines use all 316,694 Reddit side-effect relations as the reference. The 95\% Wilson intervals summarize these automatic marker shares and do not account for extraction errors.

\textbf{Adjustment for Multiple Comparisons.} We also compare each platform's PT shares with FAERS by two-sided two-proportion $z$-tests at $\alpha = 0.05$, with Bonferroni correction across the 110 tests (55 PTs on 2 platforms). Of the 46 PTs for which both narrative platforms differ from FAERS in the same direction under the Wilson interval rule, 40 differ significantly on both platforms after correction (maximum $p=0.036$). This maximum is taken across the 80 comparisons for these 40 PTs, with correction over all 110 tests.

\textbf{Robustness Check.} Figure~\ref{fig:rq2-overall}c repeats the Reddit comparison by brand. The eligibility checks below keep the Wilson interval rule. Allowing any FAERS role for study drugs gives 340,464 reports with a retained event and changes 1 of 110 classifications. Restricting Reddit to named-drug posts keeps 77,165 posts and changes 3 of 55 classifications. Restricting FAERS to consumer reports keeps 239,142 reports, 80.8\% of those compared, and changes 5 of 110 classifications.

\FloatBarrier
\vspace{-3mm}
\section{Ethical Considerations}
\label{app:ethics}
All data in this study are publicly available. We collect Reddit posts through a public archive, WebMD reviews from public drug pages, and FAERS reports through openFDA. We collect no Reddit usernames, neither the KG nor AskGLP1 keeps account names, and we make no attempt to identify or contact users. Every post shown in our figures is paraphrased. AskGLP1 shows the source sentences of each relation so that AskGLP1 users can check it, but it shows no sentences from users aged 13 to 18. Project members annotated posts and rated system answers as part of the research team, and no information about them is collected, so this work studies public posts and system outputs and is not human subjects research. The agent gives no medical advice and declines such requests with a referral to a doctor or pharmacist. It reports shares of posts that mention a side effect, which do not estimate incidence or risk, so its answers are not a basis for clinical decisions. We use MedDRA under an institutional subscription. Our repository releases code, prompts, and evaluation questions under the MIT License but no Reddit posts, WebMD reviews, or MedDRA files.

\end{document}